%% file: Main_AAAI2026.tex
\PassOptionsToPackage{table,svgnames}{xcolor}
\documentclass[letterpaper]{article} 
\usepackage{aaai2026}  
\usepackage{times}  
\usepackage{helvet}  
\usepackage{courier}  
\usepackage[hyphens]{url}  
\usepackage{graphicx} 
\usepackage{natbib}  
\usepackage{caption} 
\usepackage{algorithm}
\usepackage{algorithmic}

\usepackage{newfloat}
\usepackage{listings}
\DeclareCaptionStyle{ruled}{labelfont=normalfont,labelsep=colon,strut=off} 
\floatstyle{ruled}
\newfloat{listing}{tb}{lst}{}
\floatname{listing}{Listing}
\usepackage{algorithm}
\usepackage{algorithmic}

\usepackage{newfloat}
\usepackage{listings}
\usepackage{amsmath,amssymb,amsfonts}
\usepackage{textcomp}
\usepackage{xcolor}
\usepackage{amsthm}
\usepackage[table]{xcolor}
\usepackage{mathabx}
\usepackage{multirow}
\usepackage{float}
\usepackage{subfig}

\usepackage{array} 
\usepackage{booktabs}
\usepackage{lineno}

\usepackage{newfloat}
\usepackage{listings}

\title{Vertical Federated Continual Learning via Evolving Prototype Knowledge}
\author{
    Shuo Wang\textsuperscript{\rm 1},
    Keke~Gai\textsuperscript{\rm 1},
    Jing Yu\textsuperscript{\rm 2},
    Liehuang Zhu\textsuperscript{\rm 1},
    Qi Wu\textsuperscript{\rm 3}
}
\affiliations{
    \textsuperscript{\rm 1}The School of Cyberspace Science and Technology, Beijing Institute of Technology, Beijing 100081, China\\
    \textsuperscript{\rm 2}The School of Information Engineering, Minzu University of China\\
    \textsuperscript{\rm 3}The Australian Institute for Machine Learning (AIML), School of Computer Science, The University of Adelaide, Adelaide, SA 5005, Australia

}

\usepackage{bibentry}

\begin{document}

\maketitle


\begin{abstract}
   \input{sec/abstract} 
\end{abstract}

\begin{links}
    \link{Code}{https://anonymous.4open.science/r/V-LETO-0108/README.md}
\end{links}

\input{sec/intro}
\input{sec/relatedwork}
\input{sec/framwork}
\input{sec/experiments}
\input{sec/conclusion}
\pdfinfo{
/TemplateVersion (2026.1)
}

\bibliography{conference}

\end{document}

%% file: sec/abstract.tex
Vertical Federated Learning (VFL) has garnered significant attention as a privacy-preserving machine learning framework for sample-aligned feature federation.
However, traditional VFL approaches ignore the challenges of class continual learning and feature continual learning, resulting in catastrophic forgetting of knowledge from previous tasks.
To address the above challenges, we propose a novel vertical federated continual learning method, named \textbf{\underline{V}}ertical Federated Continual \textbf{\underline{L}}earning via \textbf{\underline{E}}volving Pro\textbf{\underline{T}}otype Kn\textbf{\underline{O}}wledge (V-LETO), which primarily facilitates the transfer of knowledge from previous tasks through the evolution of prototypes.
Specifically, we propose an evolving prototype knowledge method, enabling the global model to retain both previous and current task knowledge. 
Furthermore, we introduce a model optimization technique that mitigates the forgetting of previous task knowledge by restricting updates to specific parameters of the local model, thereby enhancing overall performance.
Extensive experiments conducted in both Class Incremental Learning (CIL) and Feature Incremental Learning (FIL) settings demonstrate that our method, V-LETO, outperforms the other state-of-the-art methods. 
For example, our method outperforms the state-of-the-art method by 10.39\% and 35.15\% for CIL and FIL tasks, respectively.


%% file: sec/intro.tex
\section{Introduction} \label{sec:intro}

\begin{figure}
    \centering
    \includegraphics[width=1.0\linewidth]{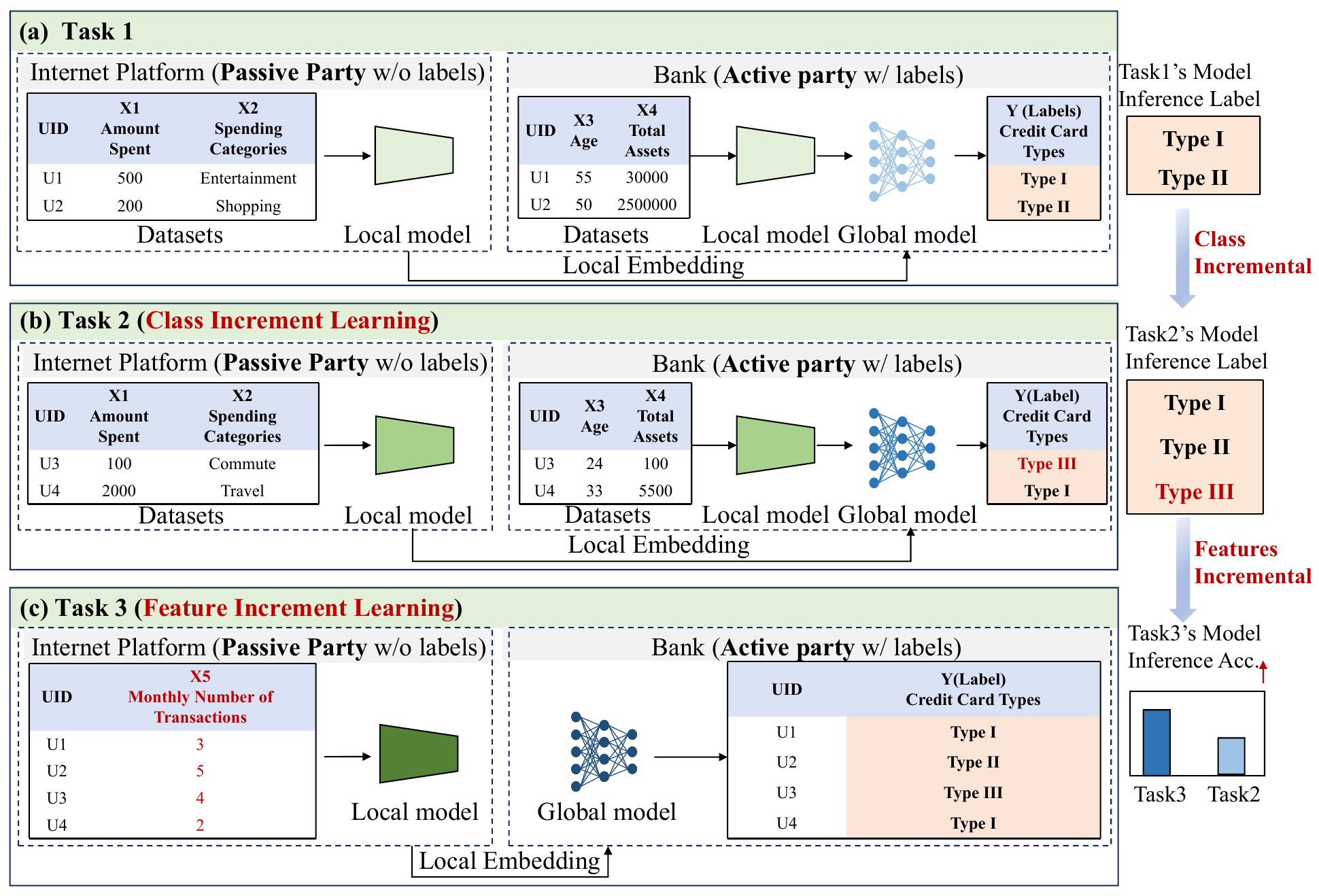}
    \caption{
    An illustration of CIL and FIL in VFL. Task 2 introduces a new class, ``Type III", for credit card marketing, while Task 3 adds a new feature, ``X5". The model's performance improves progressively from Task 1 through Task 3.
    }
    \label{fig: intro}
\end{figure}


%
 
Vertical Federated Learning (VFL)
provides multi-party collaborative computing in which datasets of different parties have overlapping samples without overlapping feature spaces \cite{castiglia2022compressed,wang2023bdvfl}, being explored in various privacy-sensitive application scenarios, e.g., financial services \cite{liu2023vertical} and healthcare \cite{sakib2024explainable}. 
In real-world applications, as users' local data increases, achieving Vertical Federated Continuous Learning (VFCL) is an expected development direction, which is generally facilitated by the integration of Class Incremental Learning (CIL) and Feature Incremental Learning (FIL).
As shown in Figure \ref{fig: intro}, take the credit card scenario for example \cite{author2024},
%
the training sample only contains ``Type I" and ``Type II" credit cards in Task 1, while Task 2 further includes user data for ``Type III" credit card. 
Therefore, after completing Task 2, the model should be able to predict all credit card types, Type I-III.
This presents a case of 
CIL, i.e., the ability of the model to learn new classes over time \cite{lebichot2024assessment,ma2022continual,casado2023ensemble}.
Another case is that the internet platform can utilize the new consumption feature ``X5" deriving from tracking the monthly consumption frequency of each user, so that the credit card marketing model derived from Task 2 can be optimized. 
This presents a case of
FIL that involves incorporating new features to enhance model performance~\cite{ni2024feature}, refer to Task 3 in Figure \ref{fig: intro}(c). 
The key to implementing VFCL lies in addressing Catastrophic Forgetting (CF) of classes and features.

However, existing VFCL supportive technologies still encounter a variety of technical challenges. 
To be specific, on one hand, most existing methods fail to address class CF of previous task knowledge in the local model of the passive party in VFL. 
Previous studies in CIL have mostly tried following strategies, including regularization, Dynamic Network Expansion (DNE), and replay. 
Among them, regularization methods adjust the learning algorithm to limit changes to key weights, preserving essential knowledge \cite{yu2024overcoming}.
Both regularization and DNE methods adjust or expand parts of the whole model to mitigate CF of previous classes \cite{luo2023gradma}.
However, the passive party only holds a partial model (ref to local model) and does not have the whole model, which includes both the local and global models.
In addition, the replay technique mitigate class CF by reconstructing previous task datasets using labels \cite{li2024towards}, but each passive party only has partial feature sets without labels in VFCL, making this technique ineffective for addressing class catastrophic forgetting.

On the other hand, the local model loses feature knowledge from previous tasks, causing feature CF issue. 
Prior studies have explored FIL in the activity recognition tasks. 
For example, FIRF \cite{8410016} is a typical method that adds nodes for newly introduced features to enhance decision tree performance. However, this method is restricted to feature augmentation in simple model structures, such as tree models, and cannot be applied to more complex model structures, such as neural network models.
Some other work tried regularization across the full model to prevent forgetting of previous features \cite{ni2024feature,10227560}. 
However, models in VFL generally is divided into local and global models.
Passive parties perform local model updates based on the gradients received from the server.
But, they only have access to the gradients of the current task and cannot observe those from previous tasks, especially when the active party employs regularization to mitigate catastrophic forgetting of previously learned features.
Therefore, retaining feature knowledge from both previous and current tasks in the local model is a key challenge in addressing feature catastrophic forgetting in VFCL.

To address the aforementioned challenges, 
we propose \underline{V}ertical Federated Continual Learning via \underline{E}volving Pro\underline{t}otype Kn\underline{o}wledge (V-LETO), a novel VFCL framework designed to overcome class- and feature-level forgetting by enabling incremental learning for both local and global models.
To addressing the loss of prior knowledge in the global model, we propose a evolving prototype knowledge method to transfer knowledge from previous tasks.
Additionally, the forgetting of prior knowledge in the local model is mitigated by constraining its updates.
Specifically, V-LETO consists of three modules. 
To address the inability of passive parties to construct prototypes due to incomplete features, we propose a prototype generation module to preserve prior task knowledge. 
The server aggregates global embeddings, derived from the passive party's local embeddings and labels, to build class prototypes.
We also propose a prototype evolving module to mitigate catastrophic forgetting of prior knowledge, which evolves prototypes that integrate both previous and current task knowledge.
Finally, we propose a model optimization module to optimize both global and local models, mitigating issues related to class- and feature-level catastrophic forgetting.

Our contributions are as follows:
(1) We propose a novel VFCL method to address an underexplored issue in VFL, which is critical in many real-world applications. To the best of our knowledge, this may be the first to attempt the simultaneous implementation of CIL and FIL in the VFL.
(2) We propose V-LETO as a framework for implementing CIL and FIL within VFL. 
To address catastrophic forgetting of prior task knowledge, we propose a global model optimization method based on evolving prototypes, which combines both prior and current task knowledge. 
Additionally, we propose a method that constrains local model updates, mitigating the catastrophic forgetting of previous task knowledge in the local model.
(3) 
We conduct extensive experiments on four datasets to evaluate the performance of the V-LETO, demonstrating its superiority over several state-of-the-art methods.
Our method outperforms the baseline method by 10.39\% and 35.15\% for CIL and FIL tasks, respectively. 
We visualized the evolving prototypes and conducted ablation and hyperparameter analysis to further evaluate our method.

%% file: sec/relatedwork.tex
\section{Related Works} \label{sec: relatedwork}

\noindent\textbf{Vertical Federated Learning.}
%
VFL enables collaborative model training across organizations that hold complementary feature sets for the same samples \cite{romanini2021pyvertical,ye2025vertical}. 
Most existing VFL approaches assume static data distributions, focusing on privacy-preserving feature alignment \cite{liao2025privacy}, model aggregation \cite{wang2025ropa}, and communication \cite{wang2025pravfed}. 
However, in real-world applications, data and environments at each participant evolve over time, making static VFL strategies insufficient. 
Moreover, current VFL methods lack mechanisms to address CF, particularly the joint class- and feature-level forgetting that arises in dynamic settings. 

\noindent\textbf{Continual Learning.}
CL seeks to mitigate CF of previously learned knowledge when models are updated on new tasks. 
Prior work can be grouped into three main categories. 
Data replay methods store a subset of past data to rehearse earlier tasks \cite{krawczyk2024analysis}; while effective, retaining raw samples raises privacy concerns. 
Parameter separation approaches isolate or expand model parameters so that new tasks do not overwrite weights important to prior tasks \cite{qiao2024learn,8410016}. 
Knowledge distillation strategies preserve prior knowledge by constraining updates through distillation or auxiliary losses \cite{li2024continual,ni2024feature}. 
However, these CL methods generally assume that central client holds complete data and a full model, and fail to address the spatial CF that arises in VFL.

\noindent\textbf{Federated Continual Learning.}
FCL addresses catastrophic forgetting across sequential tasks and distributed clients. Existing FCL methods can be divide into two main categories. 
Model parameter decompose methods preserve prior knowledge by decomposing or expanding model architectures when adapting to new tasks \cite{wang2024traceable,yoon2021federated,yang2024federated,zhang2022cross}, but they assume each client has complete labels and full models, and often introduce additional communication overhead. 
Prototype-based methods, on the other hand, replay or fuse prototype representations of previously learned tasks to mitigate forgetting without altering model structure \cite{ma2022continual,yu2024overcoming,zhu2021prototype,shenaj2023asynchronous,gao2024fedprok}.
However, these methods are designed for Horizontal FCL (HFCL) and are not directly applicable to VFCL, where clients hold only partial feature subsets and, apart from the active party, lack label access. 
Consequently, existing methods fail to address class- and feature-level catastrophic forgetting in VFCL.

%% file: sec/framwork.tex
\begin{figure*}
    \centering
    \includegraphics[width=1.0\linewidth]{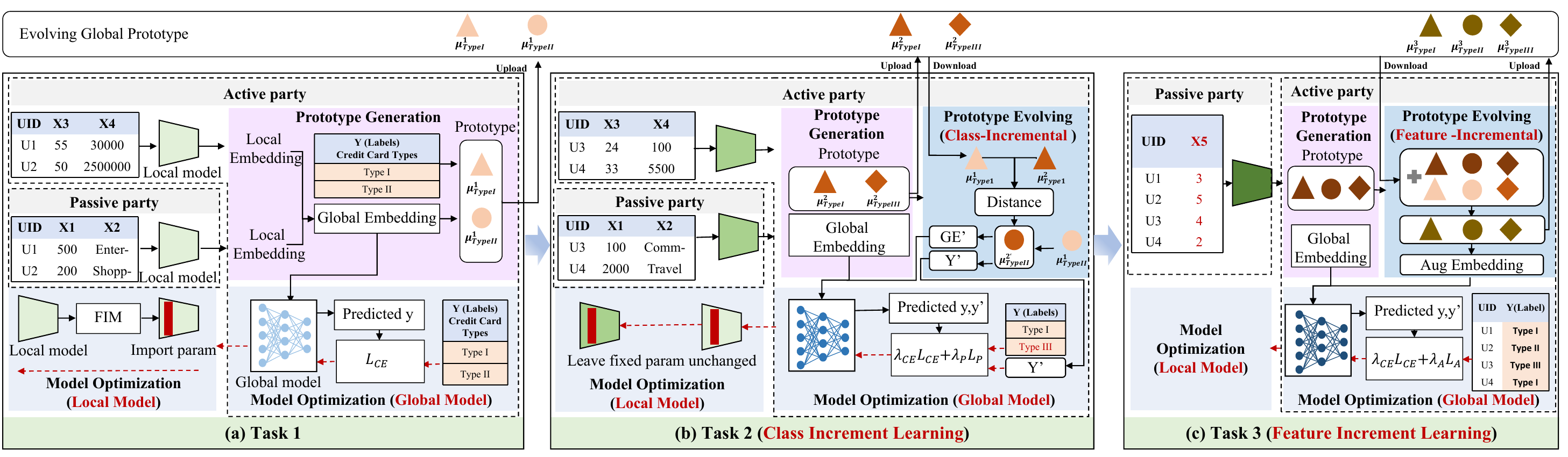}
    \caption{
    The framework overview of V-LETO. 
    V-LETO consists of three modules:
    \textit{Prototype Generation} module use local labels in conjunction with global embeddings to generate class-specific prototypes. \textit{Prototype Evolving} module uses the outputs from PG module and evolves the prototype knowledge from previous tasks for both CIL and FIL.
    \textit{Model Optimization} module optimizes the model by integrating knowledge from both previous and current tasks.
}
    \label{fig: model}
\end{figure*}

\section{Problem Definition} \label{sec: problemformulation}

Consider a typical VFL setting \cite{romanini2021pyvertical,liu2024vertical}, assume that there exists one active party $l_1$ collaborating with $K-1$ passive parties to tackle classification tasks.
For a given training task $t$, each passive party $l_k$ holds a subset of features $x_{ik}^t$ of the aligned sample data $i$ and the local model $ \mathcal{B}_k^t$. 
The active party possesses the label $Y_{i}^{t}$ of sample data $i$ and the server model $\mathcal{T}^t$. 
We assume that the active party $l_1$ and all passive parties $l_k$ have aligned training dataset samples as existing works that are obtained from the privacy set intersection \cite{luo2021feature, zhang2022adaptive}.
The objective of VFL is to collaboratively train the model between the active and passive parties to minimize the loss function, as following.
%
%
\begin{equation} \label{eq:1}
\resizebox{0.42\textwidth}{!}{$
    \min\ell(\mathcal{T}^t; \{\mathcal{B}_k^t\}_{k = 2}^K; D) \triangleq \frac{1}{N} \sum_{i = 1}^{N} \mathcal{L} \left(\mathcal{T}^t(\{E_{ik}^t\}_{k = 2}^K); Y_{i}^{t}\right)$},
\end{equation} 
%
%
where \( D = (X^t, Y^t) \) represents the training dataset; 
\( E_{ik}^t \) is the local embedding of the \( l_k \)-th passive party; 
\( X = \{\mathbf{x}_i\}_{i=1}^{N} \) and \( \mathbf{x}_k = \bigcup_{i=1}^{N} \{x_{ik}^t\} \); 
\( N \) is the total number of samples; 
$x_{ik}$ represents the local features;
\( \mathcal{L} \) is the loss function.


VFCL trains a model through multi-round collaborative learning between active and passive parties on a sequence of tasks $\{p_t \mid t = 1, 2, \dots, T\}$, where data from previous tasks becomes unavailable once a new task arrives. 
The objective is to optimize the model to minimize the loss over both current and previously encountered tasks.

\section{Methodology} \label{sec: methodology}

As shown in Figure \ref{fig: model}, V-LETO consists of three major modules: Prototype Generation (PG), Prototype Evolving (PE), and Model Optimization (MO).
Specifically, 
the PG module combines local labels with global embeddings to generate class-specific prototypes, which capture and represent the knowledge associated with each class.
PE module uses the outputs from PG module and evolves the prototype knowledge from previous tasks for both CIL and FIL.
Finally, MO module optimizes the model by integrating knowledge from both previous and current tasks. 
To mitigate the forgetting of prior task knowledge, we lock the local model parameters essential for preserving knowledge from previous tasks during updates.
We execute the modules sequentially and iterate the process until the overall model performance improves.

\subsection{Prototype Generation}
A prototype is generated by a PG module, which addresses the issue caused by fact that passive participants only have partial sample features and lack labels in VFCL, making prototype construction dramatically difficult. 
In this work, we consider the prototype a prototype belonging to a certain class.
The mechanism of PG module is that an active party aggregates local embeddings from all passive parties to obtain global embeddings, so that the knowledge of all classes are encompassed. 
We use global embeddings to generate prototypes for each class for obtaining the corresponding class prototypes. 
Specifically, an active party firstly obtains all passive parties' local embeddings $E_{ik}^{t}$, where $i$ denotes the sample, $k$ denotes the passive party $l_k$, and $t$ denotes the $t$th task.  
For instance, an active party obtains all feature embeddings $E_{i}^{t}$ of the sample $i$ in the $t$th task by aggregating local embeddings $E_{ik}^{t}$ from passive parties, expressed by $E_{i}^t = \sum_{k = 2}^{K} E_{ik}^{t}$.
In addition, an active party has the label information $Y_i^t$ for each sample $i$ and can identify the sample's class $c$. 
Consequently, the active party aggregates $E_{i}^t$ to obtain the prototypes for each class $c$. The prototype denotes $\mu_{c}^{t}=\frac{1}{\left|D_{c}\right|} \sum_{c=1}^{\left|D_{c}\right|} E_{i,c}^t$, 
where $|D_{c}|$ represents the number of samples of class $c$. 

\subsection{Prototype Evolving}
The PE module is designed to mitigate the catastrophic forgetting of previous task knowledge. 
The module facilitates the evolution of prototypes and stores the prototypes in an evolving global prototype list for task-level knowledge transfer.
Since class and feature incremental tasks require distinct approaches for PE, we separate PE module into two components, 
which are Class-Incremental PE (CI-PE) modules and Feature-Incremental PE (FI-PE) modules.

\noindent\textbf{Class-Incremental PE.}
CI-PE module performs prototype evolution by estimating distance metrics relative to existing class prototypes for CIL tasks.
We address catastrophic forgetting of previous task classes by fitting a previous class $p$ that does not appear in the current task, i.e., $p \notin \mathcal{C}^t$, but $p \in \mathcal{U}^g$.
We estimate the difference in learning ability between the previous task and current task by Equation (\ref{eq: dis}) and compute the prototype of the pseudo-prior class knowledge augmenting according to the distance difference.
We add the inter-class distance to the prototype $\mu_{p}^g$ of the previous class $p$ in the prototype $\mathcal{U}$ to obtain the approximate previous class $\hat{\mu}_{p}$, shown in Equation (\ref{eq: old_task}).
\begin{equation} \label{eq: old_task}
   \hat{\mu}_{p}^t = \mu_{p}^g + \gamma\frac{1}{|\mathcal{C}^t|} \sum_{c \in \mathcal{C}^t} \operatorname{dis}\left(\mu_{c}^{t-1}, \mu_{c}^{t}\right),
\end{equation}
where $\gamma$ denotes the hyper-parameters used as weighting factors for the knowledge distance, $\mu_{c}^{t-1}$ and $\mu_{c}^{t}$ of the new class $c$ at the $(t-1)$th task and the $t$th task, respectively.
$\operatorname{dis}$ denote the pair-wise relation with cosine similarity.
\begin{equation} \label{eq: dis}
   \operatorname{dis}\left(\mu_{c}^{t-1}, \mu_{c}^{t}\right)=\frac{\mu_{c}^{t-1} \cdot \mu_{c}^{t}}{\left\|\mu_{c}^{t-1}\right\|_{2} \times\left\|\mu_{c}^{t}\right\|_{2}}.
\end{equation}

\noindent\textbf{Feature-Incremental PE.} FI-PE aggregates global prototype with current task's class prototype to achieve cross-task feature knowledge transfer.
For an FIL task $t$, current class $c$'s prototype $\mu_{c}^t$ only contains current sample feature knowledge not previou knowledge, even though VFL require all feature knowledge for model training. 
We save the class prototype list $\mathcal{U}^g$ of previous tasks during the training process so that the feature knowledge of previous tasks can be obtained from $\mathcal{U}^g$.
A prototype aggregation mechanism is developed to weight current and previous task knowledge.
\begin{equation} \label{eq: global_proto}
    \bar{\mu}_{c}^t=\left\{\begin{array}{ll}
\mu_{c}^t & c \notin \mathcal{U}^g \\
\beta \mu_{c}^t + (1-\beta) \mu_{c}^{g} & \text {otherwise}
\end{array}\right.
\end{equation}
where $\beta$ denotes the hyper-parameters used as weighting factors for the respective prototype.
We evolve a global prototype list $\mathcal{U}^g$ using $\bar{\mu}_{c}^t$.
When class $c$ belongs to global prototype list $\mathcal{U}^g$, we exchange $\bar{\mu}_{c}^t$ for $\mu_{c}^g$ (e.g. $\mu_{c}^{g} = \mu_{c}^t$); otherwise, we insert class $c$ and prototype $\bar{\mu}_{c}^t$ into $\mathcal{U}^g$.
We reserve the global prototype list $\mathcal{U}^g$ in local memory.
$\mathcal{U}^g$ contains all classes $c$ of the current task with corresponding prototypes. 

\begin{table*}[!t]
\centering  
\setlength{\tabcolsep}{8.5pt}
\begin{tabular}{lccccc|ccccc}
\toprule
\multirow{2}{*}{Methods} & \multicolumn{5}{c|}{FMNIST}  & \multicolumn{5}{c}{CIFAR10} \\
\cmidrule(lr){2-6} \cmidrule(lr){7-11}
& $T_1$ & $T_2$ & $T_3$ & $T_4$ & AVG & $T_1$ & $T_2$ & $T_3$ & $T_4$ & AVG\\ 
\midrule
Standalone       
& \cellcolor{gray!25} 95.54& \cellcolor{gray!25} 86.96&\cellcolor{gray!25} 81.83  & \cellcolor{gray!25} 90.66 & \cellcolor{gray!25} 88.00 &
\cellcolor{gray!25} 78.73 & \cellcolor{gray!25} 55.41   & \cellcolor{gray!25} 44.55 & \cellcolor{gray!25} 55.66 & \cellcolor{gray!25}58.58 \\
Pyvertical       & 
\textbf{96.15} &  57.10    & 43.01  & 37.43 & 58.42 &
69.66 &  25.92   & 20.37&  24.49 & 35.11 \\
Pass + VFL       & 
65.66     & 65.56   & 61.24 & 58.48 & 63.48 & 
51.43     & 45.96  & 31.95 & 28.71 & 39.51 \\
FedSpace + VFL   & 
67.66     & 67.58   & 63.26 & 61.42 & 64.98 & 
51.45     & 46.06   & 31.99 & 28.81 & 39.57 \\
FedProK + VFL    & 
74.69      & 65.52   & 63.72 & 59.95 & 73.97 & 
51.45    & 44.30   & 32.18 & 30.51 & 39.61 \\
\textbf{V-LETO (Our)} & 
\underline{94.38}       &  \textbf{85.71} &   \textbf{64.04}  &  \textbf{69.13} & \textbf{76.14} & 
\textbf{74.04}      & \textbf{54.29} & \textbf{42.86}  & \textbf{33.86 } & \textbf{52.26} \\
\midrule
\midrule
\multirow{2}{*}{Methods} & \multicolumn{5}{c|}{CINIC10}  & \multicolumn{5}{c}{NEWS20} \\
\cmidrule(lr){2-6} \cmidrule(lr){7-11}
&$T_1$ & $T_2$ & $T_3$ & $T_4$ & AVG & $T_1$ & $T_2$ & $T_3$ & $T_4$ & AVG \\ 
\midrule
Standalone    & 
 \cellcolor{gray!25} 80.08    & \cellcolor{gray!25} 60.00 &  \cellcolor{gray!25} 68.35 & \cellcolor{gray!25} 84.37 & \cellcolor{gray!25} 71.20 &
\cellcolor{gray!25} 87.07   & \cellcolor{gray!25} 83.05 &   \cellcolor{gray!25} 83.60 &\cellcolor{gray!25} 83.03 & \cellcolor{gray!25} 84.19 \\
Pyvertical       & 
\textbf{79.84} & 38.35    & 36.20 &33.75 & 47.53&
84.45  &57.24 & 36.91 &21.27 & 50.73 \\
Pass + VFL       & 
58.59       & 47.08    &31.05  &30.75 & 41.86 &
65.51   & 57.55 & 39.80 &30.96 & 48.45 \\
FedSpace + VFL   & 
61.32       &44.58     &28.98  &28.50 & 40.35 &
65.88    & 55.93 &  39.30 &29.18 & 47.07 \\
FedProK + VFL    & 
78.90       & 41.66    &29.92   &26.72 & 44.30  &
\textbf{86.32}    & 54.66 &  37.65 &21.75 & 51.59\\
\textbf{V-LETO (Our)} & 
\underline{79.62} & \textbf{54.50}  &\textbf{40.65} & \textbf{34.37} & \textbf{52.28}& 
\underline{85.16} & \textbf{62.45} & \textbf{46.09}  &\textbf{32.70} & \textbf{56.60} \\
\bottomrule
\end{tabular}
\caption{Comparison of V-LETO methods with baseline methods on four datasets for CIL. The bold text represents the highest test accuracy, excluding the Standalone method. ``AVG'' represents the average test accuracy across the four tasks.}
\label{table: CIL_optimized_avg_per_dataset}
\end{table*}

\begin{table*}[!t]
\centering  
\setlength{\tabcolsep}{8.5pt}
\begin{tabular}{lccccc|ccccc}
\toprule
\multirow{2}{*}{Methods} & \multicolumn{5}{c|}{FMNIST} & \multicolumn{5}{c}{CIFAR10}  \\
\cmidrule(lr){2-6} \cmidrule(lr){7-11}
& $T_1$ & $T_2$ ($\color{blue}{\uparrow}$) & $T_3$ ($\color{blue}{\uparrow}$) & $T_4$ ($\color{blue}{\uparrow}$) & AVG & $T_1$ & $T_2$ ($\color{blue}{\uparrow}$) & $T_3$ ($\color{blue}{\uparrow}$) & $T_4$ ($\color{blue}{\uparrow}$) & AVG \\ 
\midrule

Standalone       & 
\cellcolor{gray!25} 60.32     &\cellcolor{gray!25} 73.55  &\cellcolor{gray!25} 74.01 & \cellcolor{gray!25} 61.63 & \cellcolor{gray!25} 67.38 & 
\cellcolor{gray!25} 53.40   &\cellcolor{gray!25} 59.72  &\cellcolor{gray!25} 60.90  &\cellcolor{gray!25} 52.43 & \cellcolor{gray!25} 56.61 \\
Pyvertical       &  
60.60    & 64.57  & 68.21 & 67.91 & 65.32 & 
52.19   &60.43  &61.82  & 63.09  & 59.38 \\
Pass + VFL       & 
50.11  & 62.54  & 67.38 & 66.30 & 61.58  & 
50.88   &67.67  &71.69  & 65.36 & 63.90 \\
FedSpace + VFL   & 
50.34  &59.07  & 58.55 & 40.52 & 52.12 & 
50.40  &51.43  & 58.78 &49.72 & 52.58 \\
FedProK + VFL    & 
50.67  &53.13  & 21.63 &26.18 & 37.90 & 
50.21   &46.64  & 54.10 &47.24 & 49.55 \\
\textbf{V-LETO (Our)} & 
\textbf{60.93 }     &  \textbf{88.68} &\textbf{95.06}    & \textbf{97.39} & \textbf{85.52}  & 
\textbf{53.31} & \textbf{85.12}  & \textbf{87.81} &  \textbf{88.34} & \textbf{78.65} \\
\midrule
\midrule

\multirow{2}{*}{Methods} & \multicolumn{5}{c|}{CINIC10}  & \multicolumn{5}{c}{NEWS20}  \\
\cmidrule(lr){2-6} \cmidrule(lr){7-11}
& $T_1$ & $T_2$ ($\color{blue}{\uparrow}$) & $T_3$ ($\color{blue}{\uparrow}$) & $T_4$ ($\color{blue}{\uparrow}$) & AVG & $T_1$ & $T_2$ ($\color{blue}{\uparrow}$) & $T_3$ ($\color{blue}{\uparrow}$) & $T_4$ ($\color{blue}{\uparrow}$) & AVG \\ 
\midrule

Standalone       &
\cellcolor{gray!25} 43.61& \cellcolor{gray!25} 50.86   &  \cellcolor{gray!25} 50.84   &\cellcolor{gray!25} 44.07 & \cellcolor{gray!25} 47.35  & 
\cellcolor{gray!25} 50.83   & \cellcolor{gray!25} 52.52 &  \cellcolor{gray!25} 44.16 &\cellcolor{gray!25} 51.86 & \cellcolor{gray!25} 49.84\\
Pyvertical       &
43.36    & 51.68 &53.95  & 51.60 & 50.15&
50.96   & 52.90 &  53.82 &54.44 & 53.03\\
Pass + VFL       &
36.35 & 49.07  & 53.80  & 54.55 & 48.44  &
60.32   & 70.35 &  66.99 &66.33 & 65.99\\
FedSpace + VFL   &
\textbf{46.57}  & 53.98 & 54.62  & 48.31 & 50.87 &
\textbf{66.71}    & 72.89 &  73.09 &73.13 & 71.45\\
FedProK + VFL    &44.86    & 42.89   & 41.55  &39.51 & 42.20 &
66.04   & 67.09 &  66.23 &64.36 & 65.93 \\
\textbf{V-LETO (Our)} &
41.75  & \textbf{82.93}&  \textbf{83.44}&\textbf{84.97} & \textbf{73.27} 
&
\underline{65.46} & \textbf{86.24} & \textbf{89.90} &\textbf{89.97} & \textbf{82.89} \\
\bottomrule
\end{tabular}
\caption{Comparison of V-LETO methods with baseline methods on four datasets for FIL. ``$\uparrow$'' indicates the improvement in model accuracy from the previous task to the current task. 
}
\label{table: FIL_optimized_avg_per_dataset}
\end{table*}

\subsection{Model Optimization}
This module is designed to optimize the global model and local models by using the knowledge of both previous and current task.
Two components include Gobal Model Optimization (GMO) and Local Model Optimization (LMO).


\noindent\textbf{Global Model Optimization.}
For the current task $t$, we use the global embedding obtained from the current task data to obtain the cross-entropy loss function $\mathcal{L}_{CE} = \mathcal{T}^t(E_i^t, Y_i^t)$.
For CIL and FIL tasks, we generate vectors $\bar{\mu}_{c}^t$ and $\hat{\mu}_{p}^t$ with the same batch size based on the enhanced class prototype $\hat{\mu}_{p}^t$ and the enhanced feature prototype $\bar{\mu}_{c}^t$. 
Then, we calculate the prediction value and its label based on the constructed prototype vector to obtain the loss $\mathcal{L}_A$ and $\mathcal{L}_{F}$, from $\mathcal{L}_A = \sum_{n} \mathcal{L}(\mathcal{T}^t(\hat{\mu}_{p}^t[n]), Y_p[n])$ and $\mathcal{L}_{F} = \mathcal{L}(\mathcal{T}^t(\bar{\mu}_{c}^t[n], Y_c[n]))$.
We train and optimize model parameters for obtaining the following loss, refer to Equation (\ref{eq: L}).
\begin{equation} \label{eq: L}
  \mathcal{L} = \lambda_{CE} \mathcal{L}_{CE} + \lambda_A \mathcal{L}_A + \lambda_F \mathcal{L}_{F} ,
\end{equation}
where $\lambda_F$ and $\lambda_A$ hyperparameters are weighting factors for the respective losses. 
We use the loss $\mathcal{L}$ and stochastic gradient descent to backpropagate and update the server model parameters.
The server sent the loss $\mathcal{L}$ to the passive party.

\noindent\textbf{Local Model Optimization.}
The passive party computes the gradient value \(\textit{g} = \frac{\partial \mathcal{L}}{\partial E_k}\) and updates the local model using the loss value \(\mathcal{L}\) provided by the active party.
However, the gradient \( g \) contains only the knowledge of the current task and lacks information from previous tasks. To enable the passive party to retain prior knowledge during model training, we have adopted a method that involves fixing the parameters critical to previous tasks, thereby preserving those model parameters that are particularly significant to the prior tasks.
Specifically, we estimate the importance of each parameter to the previous tasks by calculating the Fisher Information Matrix (FIM) \cite{yang2023dynamic}.
\begin{equation} \label{eq: AF}
    \mathcal{F}_{ki} \approx \frac{1}{N} \sum_{i=1}^{N}\left(\nabla_{\theta_{k}} \mathcal{L}\left(x_{i}, y_{i}\right)\right)^{2},
\end{equation}
where $\mathcal{F}_{ki}$ denote the FIM of local model $\mathcal{B}$ with the previous tasks, $k$ denote the $k$th passive party.

In addition, we set a threshold $\kappa$ by calculating the mean and standard deviation of FIM to select model parameters.
\begin{equation} \label{eq: thre}
    \kappa = \frac{1}{N} \sum_{i=1}^N \mathcal{F}_{ki} - \delta \sqrt{\frac{1}{N} \sum_{i=1}^N (\mathcal{F}_{ki} - \frac{1}{N} \sum_{i=1}^N \mathcal{F}_{ki})^2},
\end{equation}
where $\delta = k_0 + \alpha \log(t + 1)$ is a hyperparameter; a larger \(\delta\) value increases the emphasis on previous tasks.

We select the relatively important parameters $\mathcal{B}_{ki}^t$ for previous tasks based on a defined threshold $\kappa$. 
The important parameters are then fixed to prevent interference from new tasks, thus preserving the foundational model’s retention of knowledge from prior tasks and mitigating forgetting. 
\begin{equation} \label{eq: local_update}
    \mathcal{B}_{ki}^t=\left\{\begin{array}{ll}
\mathcal{B}_{ki'}^{t-1} & \mathcal{F}_{ki} \ge \kappa \\
\text{Update}~\mathcal{B}_{ki}^{t-1} & \text {otherwise}.
\end{array}\right.
\end{equation}

%% file: sec/experiments.tex
\section{Experiments} \label{sec: exper}



\noindent\textbf{Datasets.}
We conducted comprehensive experiments on four datasets widely used for the VFL \cite{wang2024unified}, which are 
FashionMNIST (FMNIST) \cite{xiao2017fashion}, CIFAR10 \cite{abouelnaga2016cifar}, CINIC10 \cite{NEURIPS2021_2f2b2656} and NEWS20 \cite{lang1995newsweeder}.
Following \cite{qiu2024integer}, we vertically partitioned the dataset into 
$K$ subsets, each allocated to a passive party.
For the image datasets, we employed a CNN model consisting of three convolutional layers and three fully connected layers, while an MLP4 model was used for the NEWS20 dataset.

\noindent\textbf{Baselines.}
To the best of our knowledge, most existing schemes have focused on HFCL, while VFCL has received limited attention.
However, the evaluation methodologies for these two types of FCL differ due to their distinct training mechanisms.
To ensure fair comparison, we adapted state-of-the-art methods designed for HFCL to VFCL and used them as baselines to evaluate the performance of our proposed approach, V-LETO.
The implemented baselines in our evaluations included Standalone, PyVertical \cite{romanini2021pyvertical}, Pass \cite{zhu2021prototype} + VFL, FedSpace \cite{shenaj2023asynchronous} + VFL, and FedProK \cite{gao2024fedprok} + VFL.
We compared the CIL and FIL performance of our proposed method with these baseline methods. 
Additionally, we employed independent model training for distinct tasks and a full-feature model training as the baseline method to evaluate the effectiveness of our approach. 

\begin{table*}[!t]
\centering
\setlength{\tabcolsep}{2.5pt}
\begin{tabular}{ccc|c|c|ccc|cccc|ccccc}
\toprule
\multirow{2}{*}{$\mathcal{L}_{CE}$} & \multirow{2}{*}{$\mathcal{L}_{A}$} & \multirow{2}{*}{LMO}&Training Task& $T_1$& \multicolumn{3}{c|} {$T_2$} & \multicolumn{4}{c|}{$T_3$} & \multicolumn{5}{c}{$T_4$}\\ \cmidrule{4-17}
& & & Testing Task & $T_1$ & $T_1$ & $T_2$ & $T_{12}$ & $T_1$ & $T_2$ & $T_3$ & $T_{123}$& $T_1$ & $T_2$ & $T_3$ & $T_4$ & $T_{1234}$\\
\midrule
$\times$ & $\checkmark$ & $\checkmark$ &V-LETO w/o $\mathcal{L}_{CE}$ &  93.82 & 63.54&13.59&41.87 & 93.20&61.95&16.01&57.50 & 78.59&75.53&39.92&0.00&47.57 \\
$\checkmark$ & $\times$ & $\checkmark$ &V-LETO w/o $\mathcal{L}_{A}$ &  93.82 &41.64&85.62&57.96 & 0.00&42.26&85.40&44.21 & 0.00&0.00&46.40&91.87&36.25 \\
$\checkmark$ & $\checkmark$ & $\times$ &V-LETO w/o LMO & 93.90  & 43.12&91.32&62.61 & 0.00&47.42&92.74&42.03 & 0.00&0.00&47.65&95.93&37.96 \\
$\checkmark$ & $\checkmark$ & $\checkmark$ &V-LETO (Ours)     & \textbf{94.38}  & 83.73&83.92&\textbf{85.71} & 48.82&62.18&81.17&\textbf{64.04} & 46.50&45.70&64.37&86.87&\textbf{69.13} \\
\bottomrule
\end{tabular}
\caption{Ablation study of CIL in V-LETO on FMNIST,   ``w/o'' denotes ``without''. 
When training task $T_2$, we evaluate the previous task $T_1$, the current task $T_2$, and all task $T_{12}$, where Task $T_{12}$ means using the data of Task1 and Task2 to evaluate the performance of the model.}
\label{table: ablation-class}
\end{table*}

\begin{table*}[!t]
\centering

\setlength{\tabcolsep}{4.0pt}
\begin{tabular}{cc|c|c|ccc|cccc|ccccc}
\toprule
\multirow{2}{*}{$\lambda_{CE}$} & \multirow{2}{*}{$\lambda_{A}$} &Training Task & $T_1$ & \multicolumn{3}{c|} {$T_2$} & \multicolumn{4}{c|}{$T_3$} & \multicolumn{5}{c}{$T_4$}\\ \cmidrule{3-16}
& & Testing Task & $T_1$ & $T_1$(${\color{blue}\uparrow}$) & $T_2$(${\color{blue}\downarrow}$) & $T_{12}$ & $T_1$(${\color{blue}\uparrow}$) & $T_2$(${\color{blue}\uparrow}$) & $T_3$(${\color{blue}\downarrow}$) & $T_{123}$& $T_1$(${\color{blue}\uparrow}$) & $T_2$(${\color{blue}\uparrow}$) & $T_3$(${\color{blue}\uparrow}$) & $T_4$(${\color{blue}\downarrow}$) & $T_{1234}$\\ 
\midrule
$0.8$ & $0.2$ &V-LETO&94.41  & 84.37&84.78& 84.92&40.10 &59.07& 84.90&62.67 & 37.98&31.05&59.80&91.59&57.18 \\
$0.7$ & $0.3$ &V-LETO&  94.37 & 83.51&84.35&85.19 &42.21 &59.51&84.10&62.80
 &38.89&33.05&60.64&91.05&58.14 \\
$0.5$ & $0.5$ &V-LETO&94.38  & 83.73&83.92&85.71 & 48.82&62.18&81.17&64.04 & 46.50&45.70&64.37&86.87&69.13\\
$0.3$ & $0.7$ &V-LETO & 94.41 &84.39 &81.84& 84.10&58.71 &62.19 &75.06 &66.37 &45.70 &43.22&62.13 &85.27 & 61.03 
 \\
$0.2$ & $0.8$ &V-LETO& 94.41 &84.77 &80.43 &83.39 &71.45 &63.07 &68.65 &69.86 &51.02 &47.58 &63.59 &81.50 &62.18 
\\
\bottomrule
\end{tabular}
\caption{The impact of hyperparameters $\lambda_{CE}$ and $\lambda_{A}$ on CIL model performance on the FMNIST dataset.}
\label{table: hyper_CIL1}
\end{table*}

\begin{table}[!t]
\centering
\setlength{\tabcolsep}{1.25pt}
\begin{tabular}{ccc|c|ccccc}
\toprule
 $\mathcal{L}_{CE}$ & $\mathcal{L}_{F}$ & LMO & \textbf{Settings} & $T_1$ & $T_2$ & $T_3$ & $T_4$ \\
\midrule
$\times$ & $\checkmark$ & $\checkmark$ & V-LETO w/o $\mathcal{L}_{CE}$ & 60.90 & 79.99 & 89.97 & 90.01 \\
$\checkmark$ & $\times$ & $\checkmark$ & V-LETO w/o $\mathcal{L}_{F}$ & 61.01 & 74.83 & 71.94 & 60.70 \\
$\checkmark$ & $\checkmark$ & $\times$ &V-LETO w/o LMO &  60.17 & 60.26 & 64.32 & 43.10 \\
$\checkmark$ & $\checkmark$ & $\checkmark$ & V-LETO (Ours)     &  \textbf{60.93} & \textbf{88.68}& \textbf{95.06} & \textbf{97.39} \\
\bottomrule
\end{tabular}
\caption{Ablation study of FIL on the FMNIST dataset.}
\label{table: ablation-feature}
\end{table}

\noindent\textbf{Implementation Details.}
Evaluations were performed in Python on PyTorch, using a server with an NVIDIA GeForce RTX 3090 GPU and CUDA 12.5.
The number of total tasks was configured to 4 and each task contains a set of new classes or features; the number of passive participants was configured to 4 and each participant holds a subset of features. 
We adopted a stochastic gradient descent as the optimizer of modal training with a learning rate of 1e-3.
The hyperparameters $\beta$, $\lambda_F$, and $\lambda_A$ were all set to 0.5; $k_0$ was set to 15; $\alpha$ was set to 3. 
To achieve a fair comparison, all experiments were conducted under the same configuration to evaluate the performance of different methods.




\subsection{Comparison with State-of-the-Art Methods}
\noindent\textbf{Class Incremental Learning.}
As shown in Table \ref{table: CIL_optimized_avg_per_dataset}, our approach demonstrates superior model performance in CIL.
The standalone method denotes the accuracy obtained by training each task independently without CL, resulting in relatively high performance.
However, as the number of tasks increases, baseline models gradually decline in accuracy due to class CF. 
After training on four tasks, our approach performs substantial accuracy gains. 
Additionally, its average accuracy improvement of 32.27\% demonstrates its effectiveness in address class CF of VFCL.
\begin{table*}[!t]
\centering
\setlength{\tabcolsep}{4.0pt}
\begin{tabular}{cc|c|c|ccc|cccc|ccccc}
\toprule
\multirow{2}{*}{$k_0$} & \multirow{2}{*}{$\alpha$} &Training Task& $T_1$ & \multicolumn{3}{c|} {$T_2$} & \multicolumn{4}{c|}{$T_3$} & \multicolumn{5}{c}{$T_4$} \\ \cmidrule{3-16}
& & Testing Task & $T_1$ & $T_1$ & $T_2$ & $T_{12}$ & $T_1$ & $T_2$ & $T_3$ & $T_{123}$ & $T_1$ & $T_2$ & $T_3$ & $T_4$ & $T_{1234}$\\
\midrule
$5$ & $1$ &V-LETO &94.39   & 83.12 &83.98 &84.96 & 47.11 &61.07 &81.52 &63.79 & 41.13 &38.07 &62.99 &89.61 &59.77 
  \\
$10$ & $2$ &V-LETO &94.43 &83.35 &83.82 &84.89 &47.36 &61.40 &81.21 &63.88 &41.02 &38.26 &63.30 &89.43 & 59.85 
 \\
$15$ & $3$ &V-LETO &94.38  & 83.73 & 83.92 & 85.71 & 48.82&62.18&81.17&64.04 & 46.50&45.70&64.37&86.87&69.13\\
$20$ & $4$ &V-LETO &94.39 &83.52 & 83.18 & 84.87 &46.97 &60.86 &81.09 &63.71 &40.88 &37.91 &62.81 &89.57 &59.68 
  \\
$25$ & $5$ &V-LETO & 94.39 &83.54  &83.20&84.90 &46.93 &60.82 &81.15 &63.75 &40.88 &37.97 &63.05 &89.82 &59.76   \\
\bottomrule
\end{tabular}
\caption{The impact of hyperparameters $k_0$ and $\alpha$ on CIL model performance on the FMNIST dataset.}
\label{table: hyper_CIL2}
\end{table*}

\noindent\textbf{Feature Incremental Learning.}
Table \ref{table: FIL_optimized_avg_per_dataset} presents comparisons of model test accuracy in FIL tasks. 
It depicts that V-LETO continuously improve model accuracy when the number of FIL tasks increases, greatly superior to baseline methods. 
For example, our method achieves a 35.15\% improvement in model test accuracy in task4 on the CIFAR10 datasets compared to the State-of-Art methods.
It implies that the State-of-Art methods suffer from feature forgetting and can hardly use features from previous tasks, while our method has been evidenced that it can effectively utilize features from previous tasks to optimize model parameters and improve performance.

\begin{figure}[!t] 
	\begin{minipage}{0.32\linewidth}
\centerline{\includegraphics[width=\textwidth]{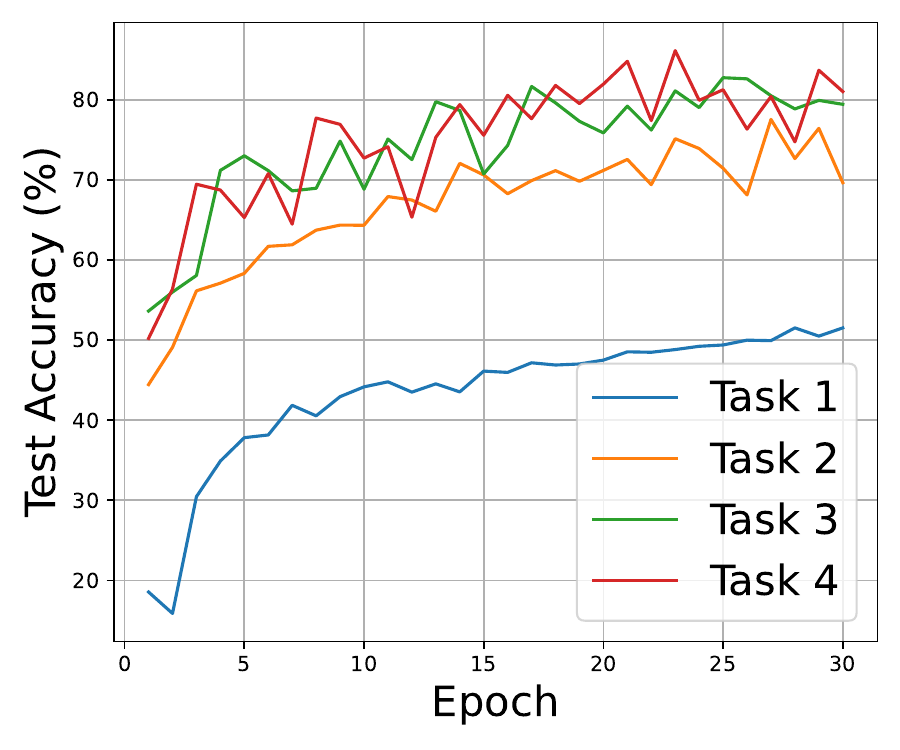}}
		\centerline{(a) (0.8 0.2)}
	\end{minipage}
    \begin{minipage}{0.32\linewidth}
\centerline{\includegraphics[width=\textwidth]{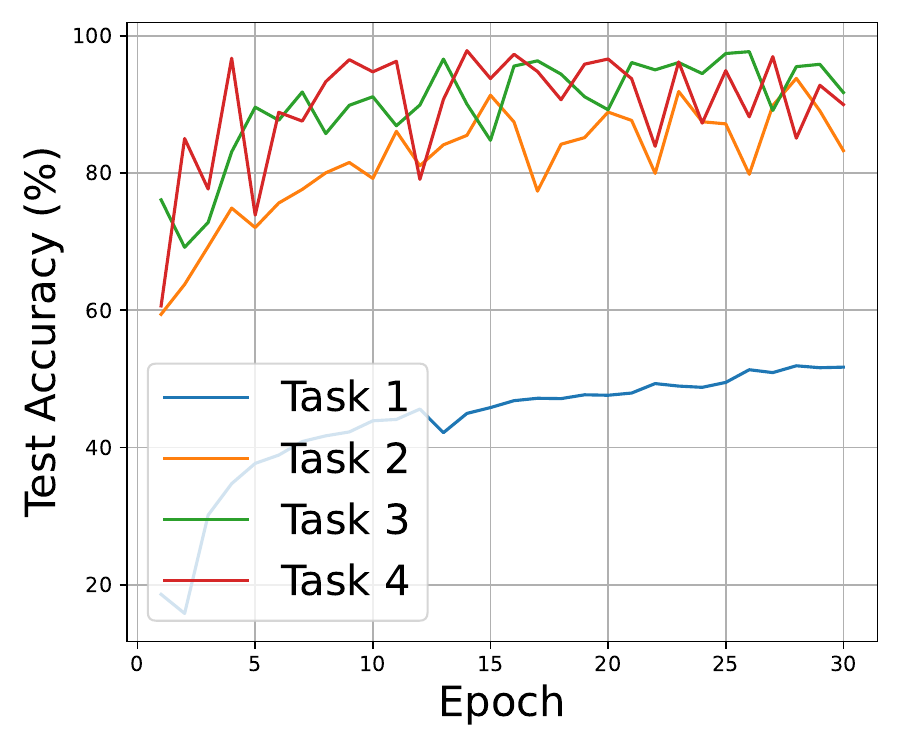}}
		\centerline{(b) (0.5 0.5)}
	\end{minipage}
	\begin{minipage}{0.32\linewidth} \centerline{\includegraphics[width=\textwidth]{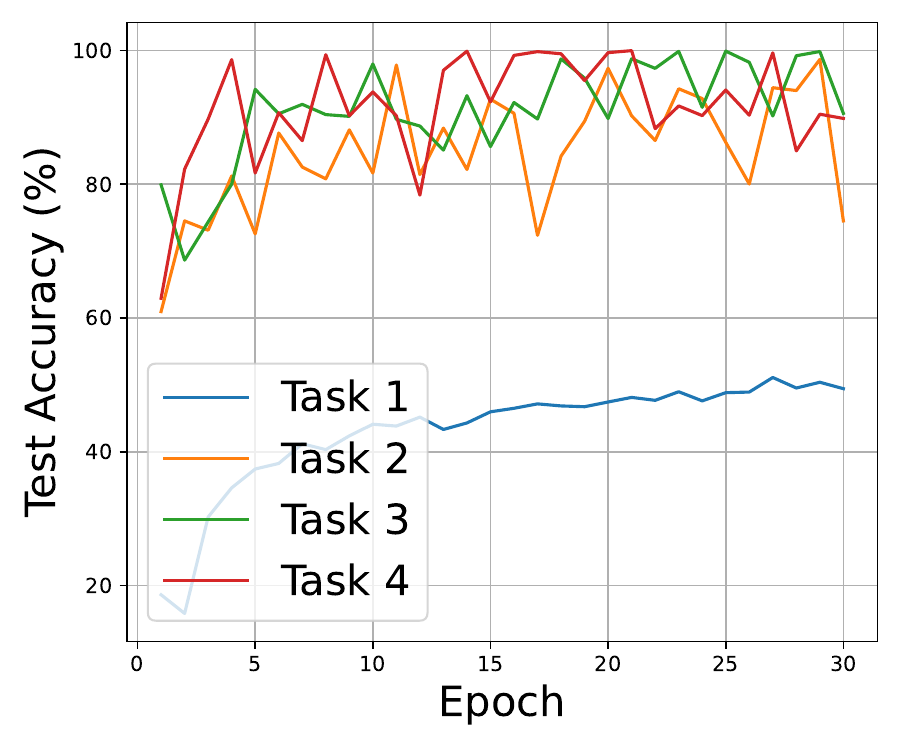}}
		\centerline{(c) (0.2 0.8)}
	\end{minipage}
	\caption{The impact of hyperparameters ($\lambda_{CE}$, $\lambda_{F}$) on FIL on the FMNIST dataset.}
	\label{Fig: hyperparameter}
\end{figure}

\begin{figure*}[!t] 
	\begin{minipage}{0.24\linewidth}
		\centerline{\includegraphics[width=\textwidth]{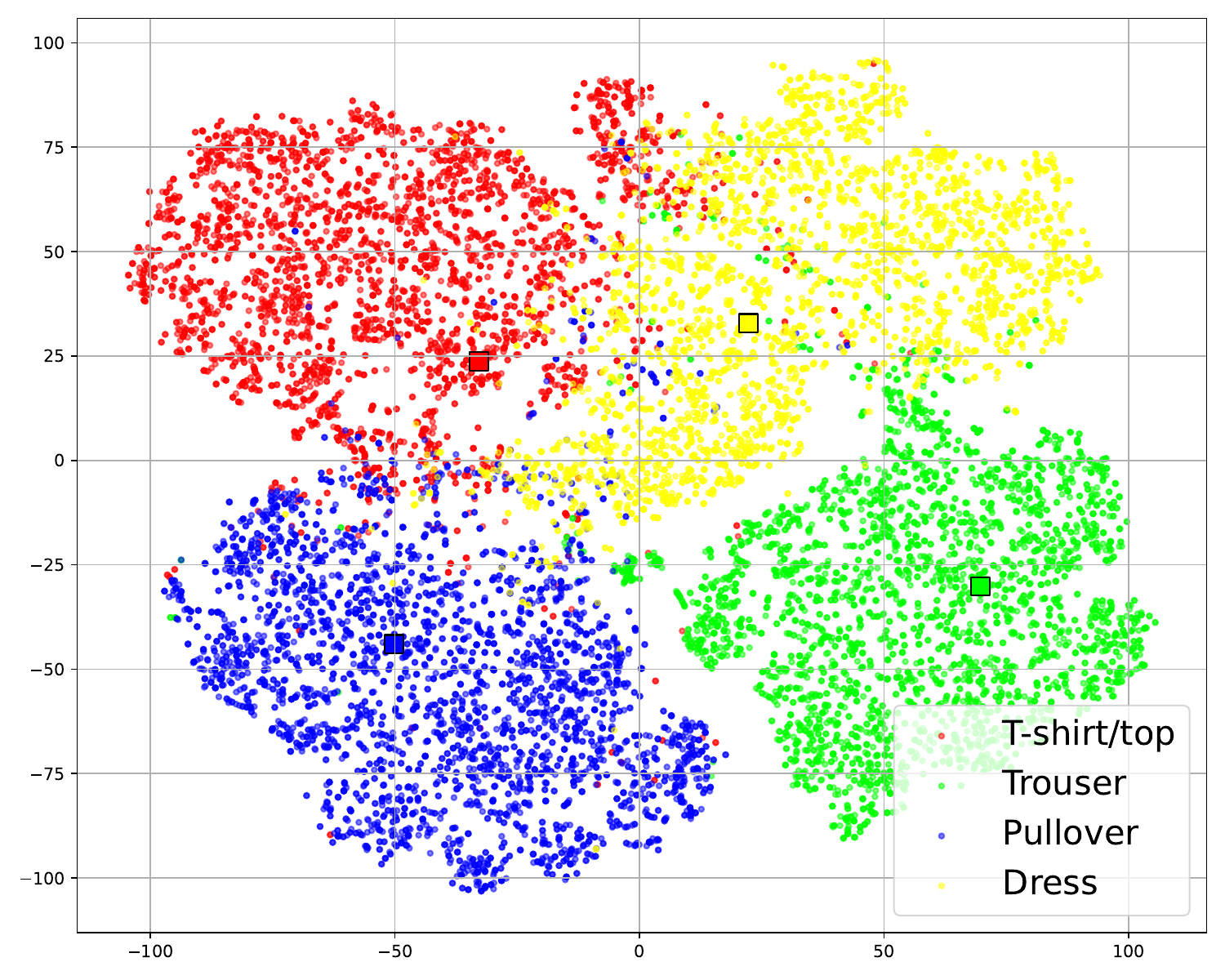}}
		\centerline{(a) $T_1$}
	\end{minipage}
    \begin{minipage}{0.24\linewidth}
		\centerline{\includegraphics[width=\textwidth]{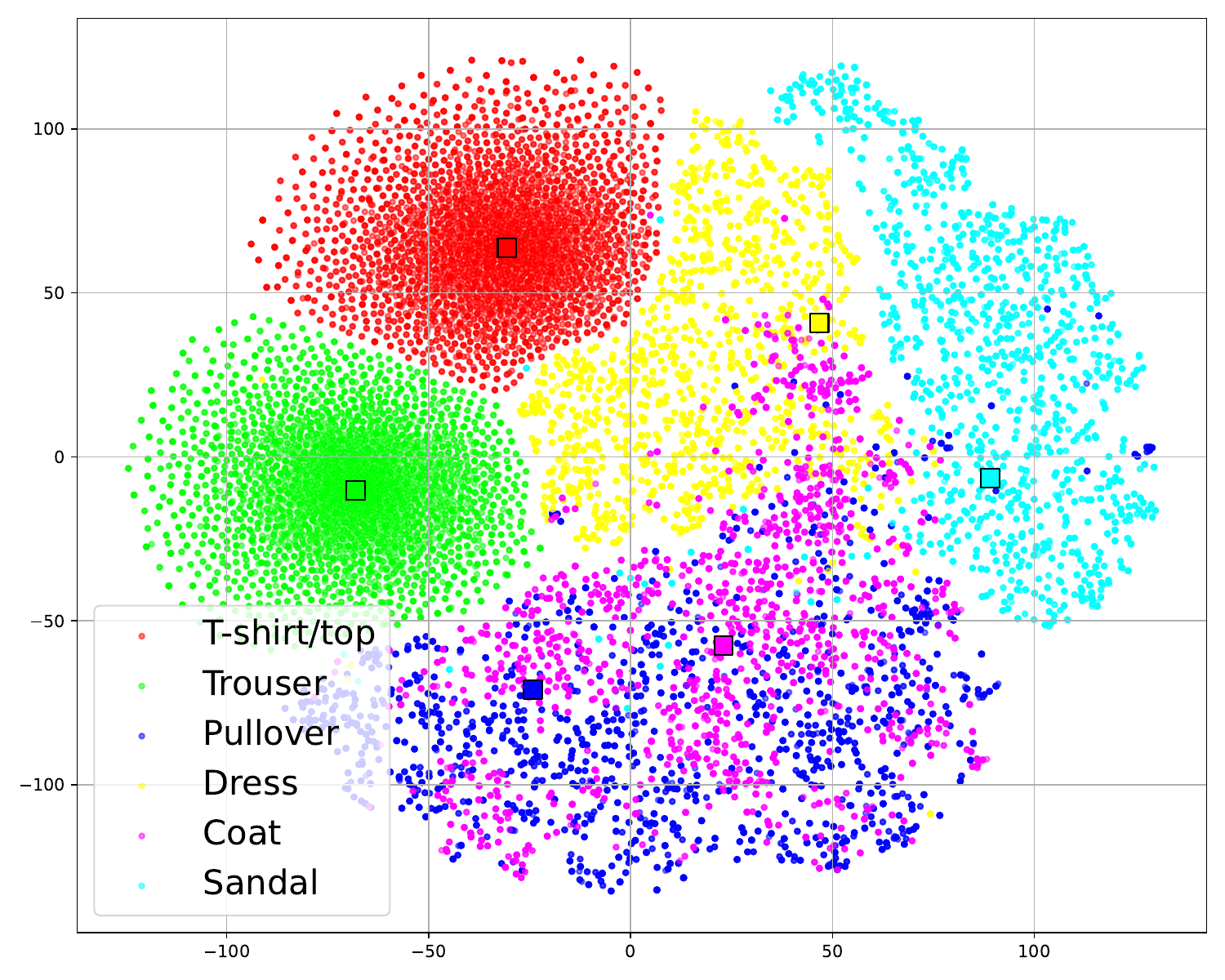}}
		\centerline{(b) $T_2$}
	\end{minipage}
	\begin{minipage}{0.24\linewidth}
		\centerline{\includegraphics[width=\textwidth]{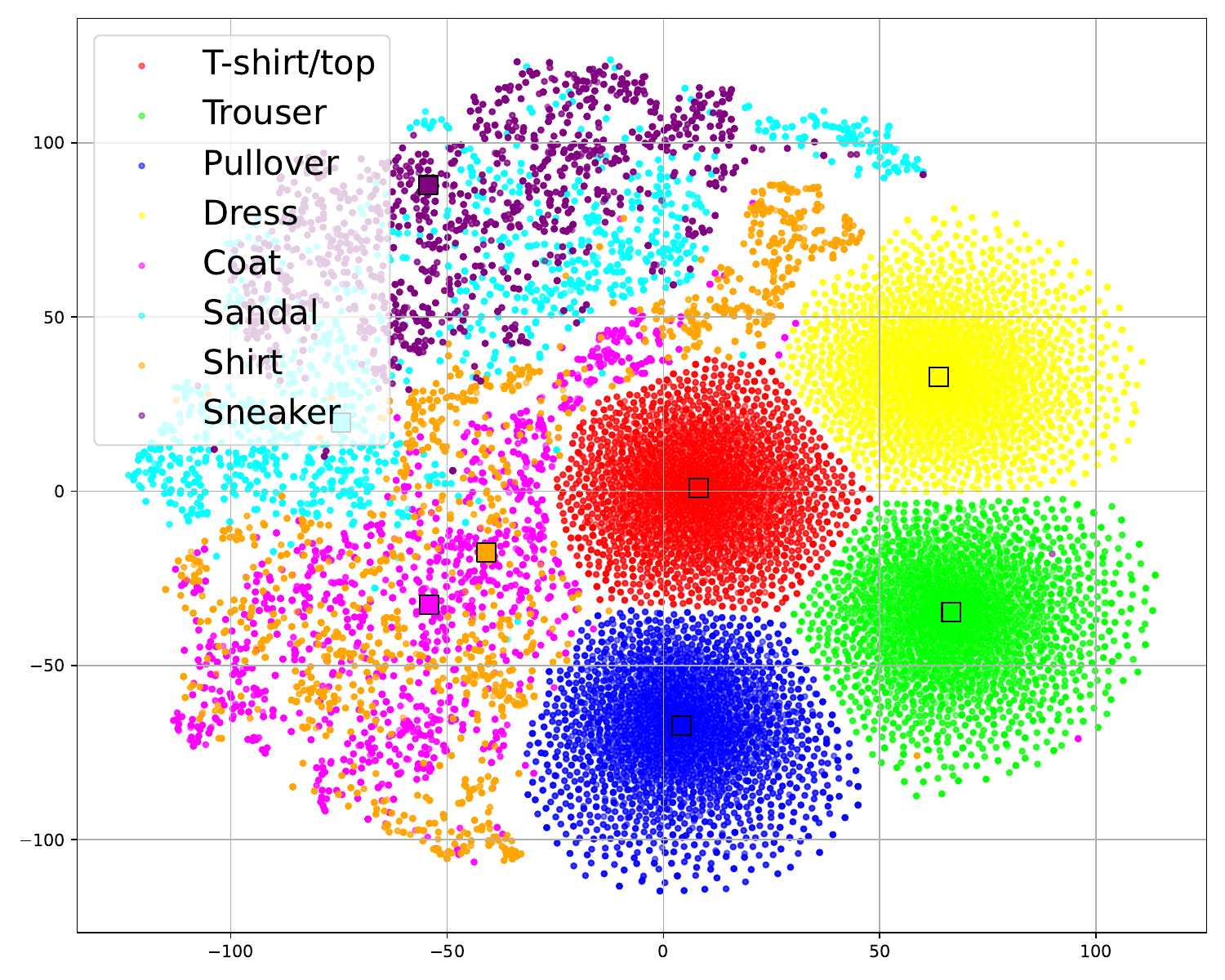}}
		\centerline{(c) $T_3$}
	\end{minipage}
    \begin{minipage}{0.24\linewidth}
		\centerline{\includegraphics[width=\textwidth]{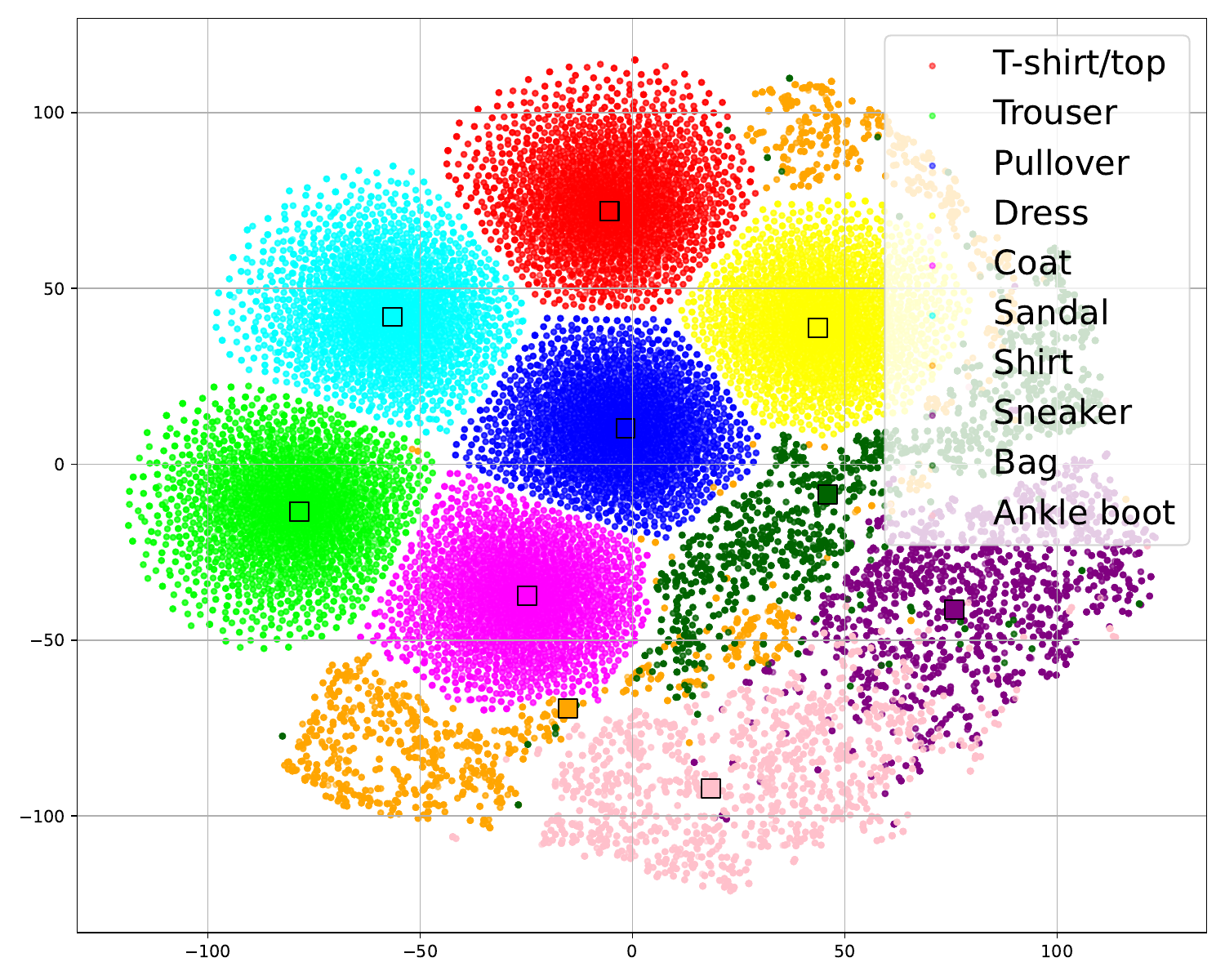}}
		\centerline{(d) $T_4$}
	\end{minipage}

    \begin{minipage}{0.24\linewidth}
		\centerline{\includegraphics[width=\textwidth]{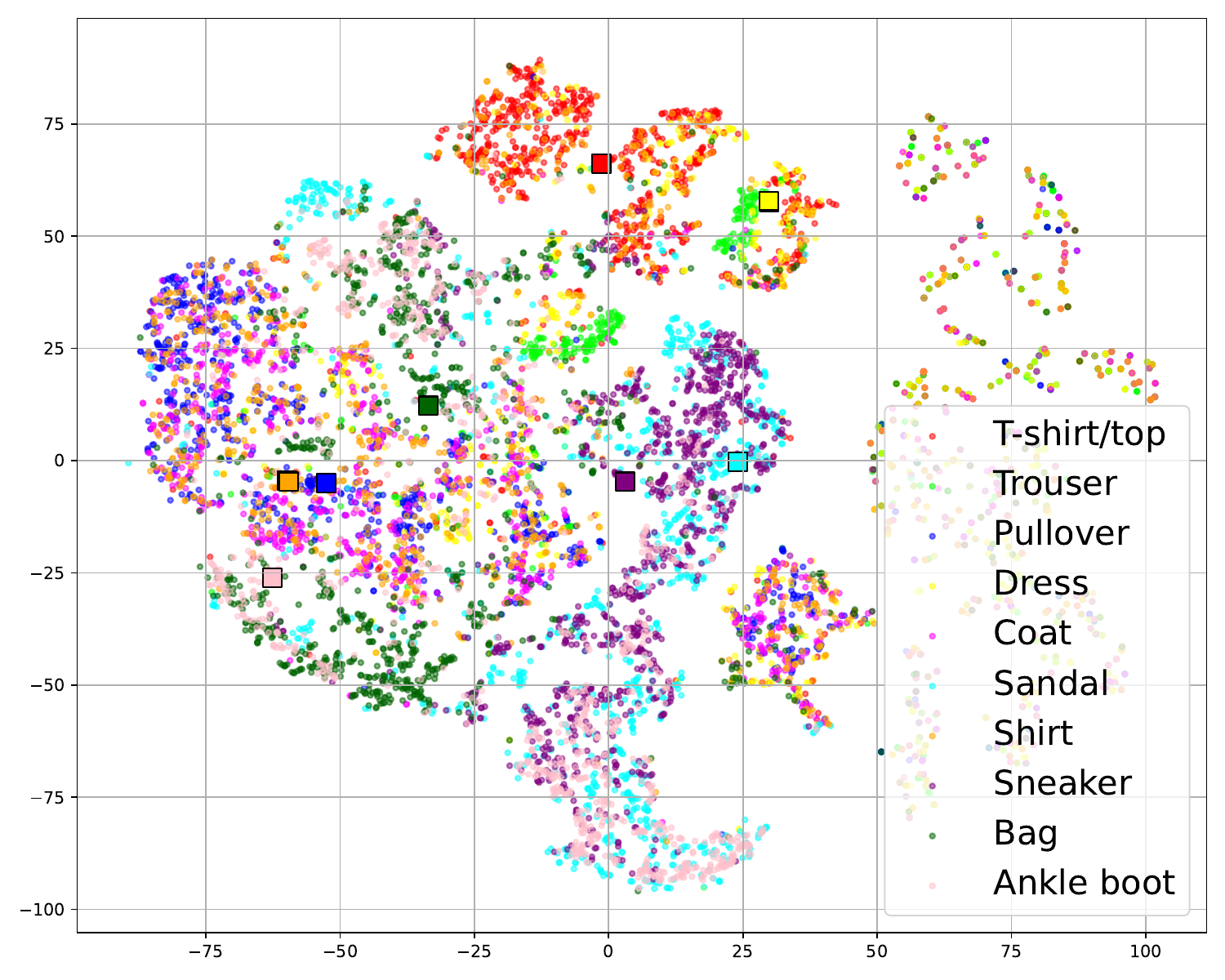}}
		\centerline{(e) $T_1$}
	\end{minipage}
    \begin{minipage}{0.24\linewidth}
		\centerline{\includegraphics[width=\textwidth]{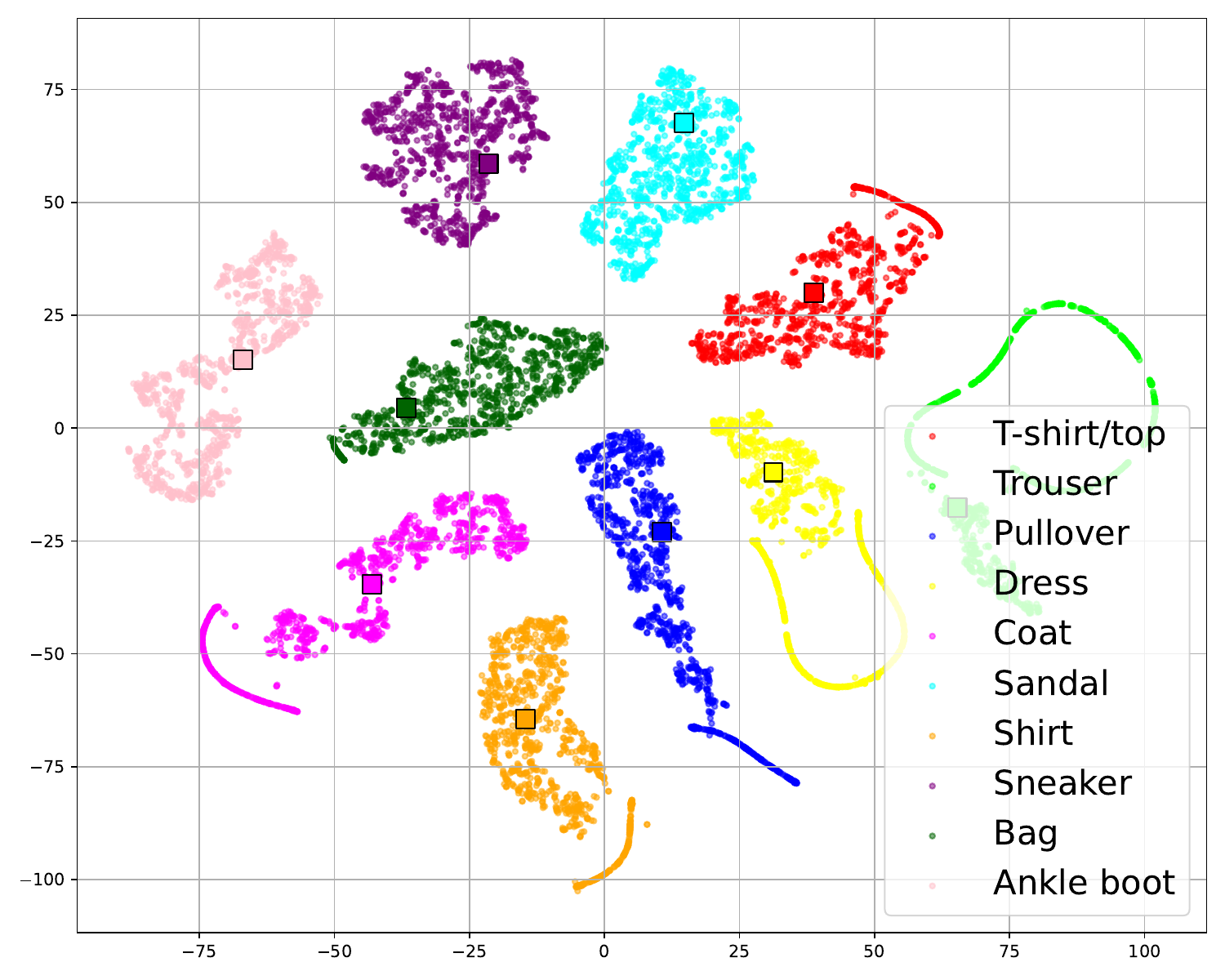}}
		\centerline{(f) $T_2$}
	\end{minipage}
	\begin{minipage}{0.24\linewidth}
		\centerline{\includegraphics[width=\textwidth]{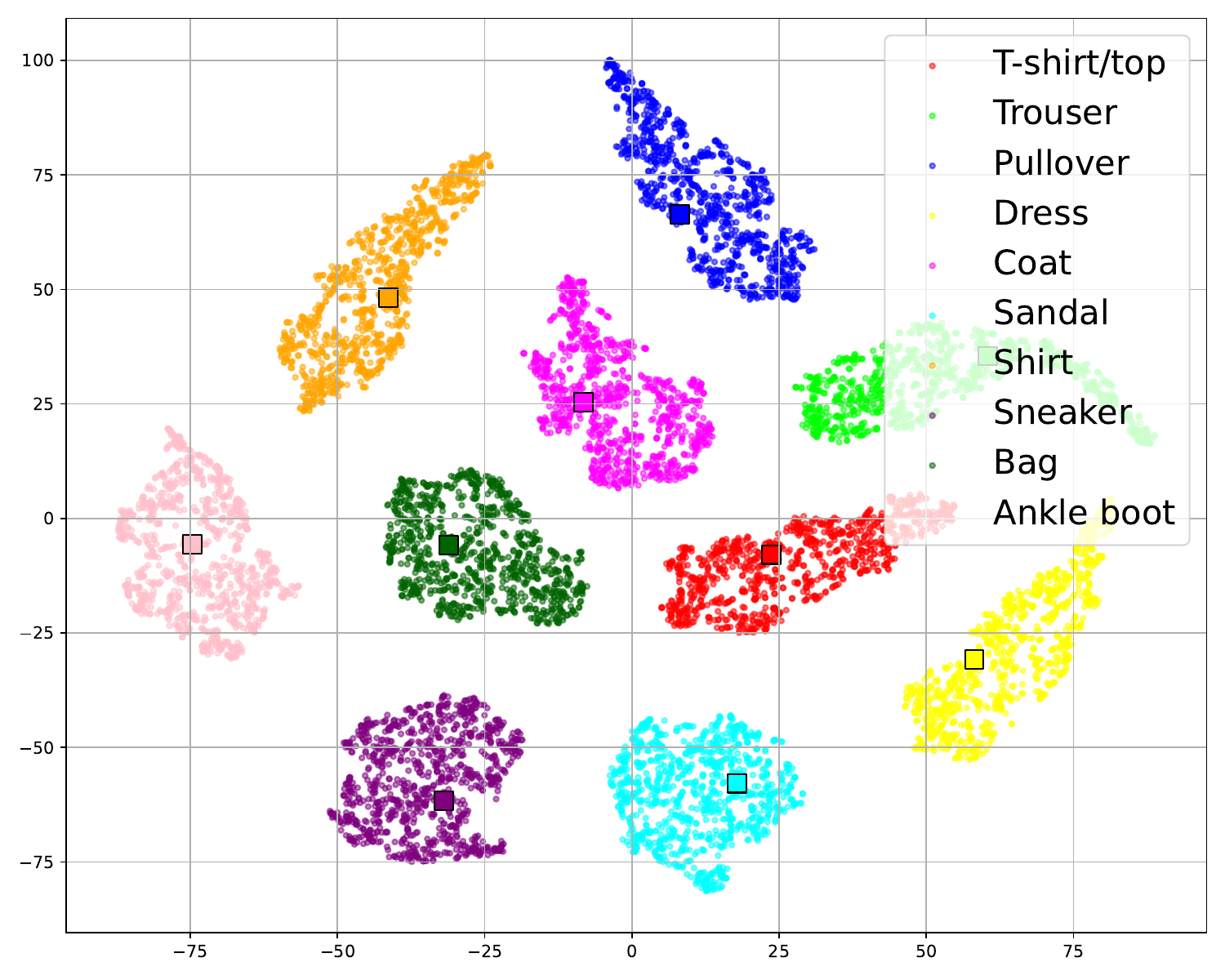}}
		\centerline{(g) $T_3$}
	\end{minipage}
    \begin{minipage}{0.24\linewidth}
		\centerline{\includegraphics[width=\textwidth]{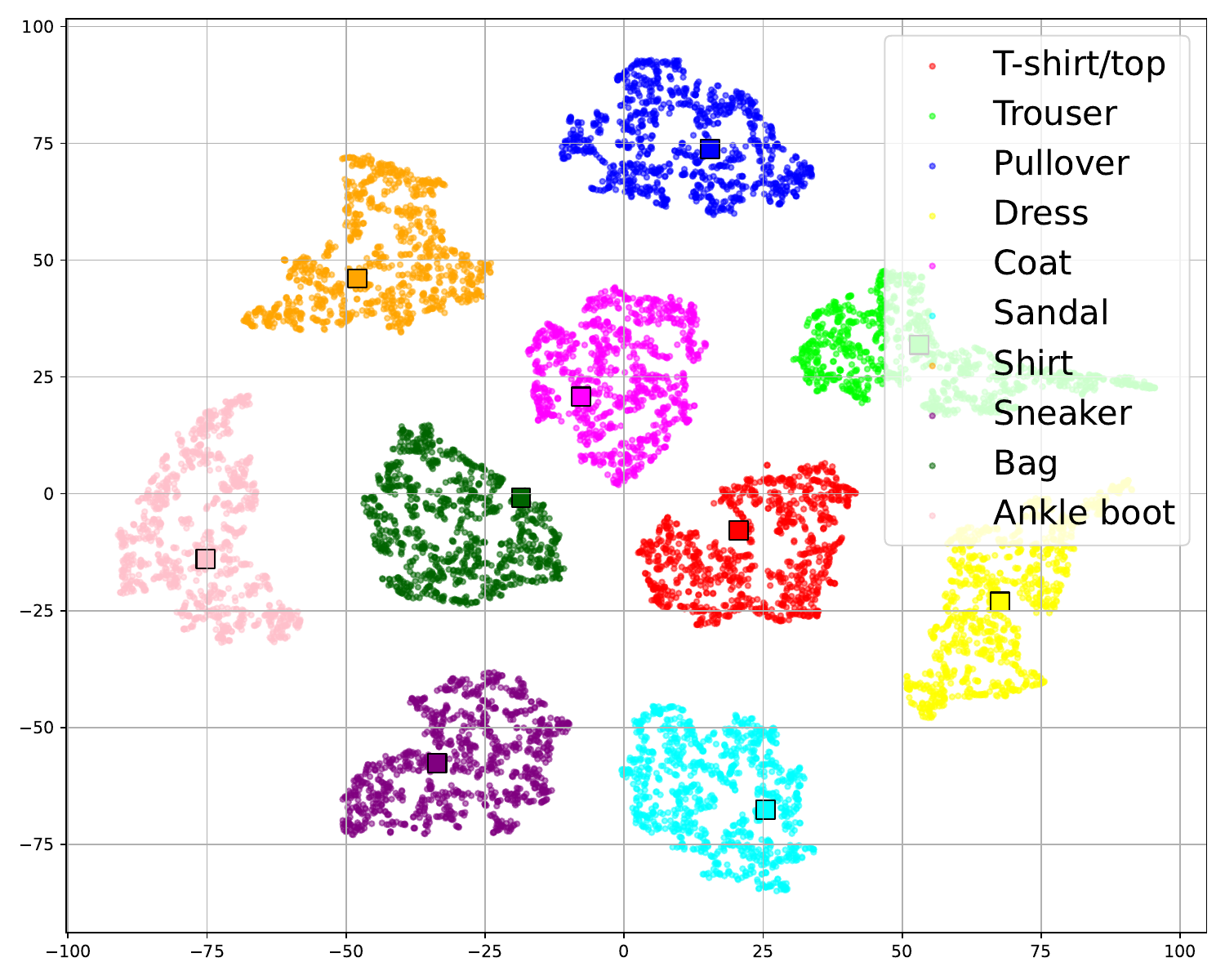}}
		\centerline{(h) $T_4$}
	\end{minipage}
\caption{$t$-SNE visualization of the global embedding and prototypes produced by CIL (a)(b)(c)(d) and FIL (e)(f)(g)(h) in V-LETO on the FMNIST dataset.
Colored circles represent global embeddings, and black squares denote class prototypes.}
	\label{Fig: Vis_CIL}
\end{figure*}

\subsection{Ablation Study}
For CIL in V-LETO, Table \ref{table: ablation-class} depicts that the performance of the current training task significantly decreases in V-LETO w/o $\mathcal{L}_{CE}$, indicating that $\mathcal{L}_{CE}$ successfully facilitates learning for the current task. 
The performance of previous tasks drops significantly in V-LETO w/o $\mathcal{L}_{A}$ and V-LETO w/o LMO, indicating that $\mathcal{L}_{A}$ and LMO effectively mitigate the class catastrophic forgetting problem and enable CIL through class prototype evolution.
In addition, for FIL in V-LETO, Table \ref{table: ablation-feature} depicts that V-LETO w/o $\mathcal{L}_{F}$ and V-LETO w/o LMO do not exhibit improved model performance as feature tasks increase, suggesting that $\mathcal{L}_{F}$ and LMO facilitate feature knowledge transfer from previous tasks, thereby enhancing model performance. 


\subsection{Hyper-parameter Analysis}
We evaluated the core hyperparameters in V-LETO, including the current task update rate ($\lambda_{CE}$), catastrophic forgetting mitigation parameters ($\lambda_{A}$, $\lambda_{F}$), and local model update parameters ($k_0$, $\alpha$).
The results of the hyperparameter analysis are presented in Tables \ref{table: hyper_CIL1}, \ref{table: hyper_CIL2} and Figure \ref{Fig: hyperparameter}.
Table \ref{table: hyper_CIL1} depicts that the accuracy of previous tasks continuously increases while the accuracy of the current task declines, as $\lambda_{CE}$ decreases and $\lambda_{A}$ increases.
Table \ref{table: hyper_CIL2} shows that the importance of old task parameters in the local model increases as $k_0$ and $\alpha$ increase, with optimal model performance achieved when $k_0 = 15$ and $\alpha = 5$. 
Figure \ref{Fig: hyperparameter} exhibits that FIL task model performance is optimized when $\lambda_{CE} = 0.5$ and $\lambda_{F} = 0.5$.
These findings demonstrate that hyperparameter optimization is crucial for achieving optimal model performance.


\subsection{Effectiveness of Prototype Evolving}

We apply $t$-SNE \cite{cai2022theoretical} to visualize the samples in the FMNIST dataset. 
Figures \ref{Fig: Vis_CIL}(a)-(d) show visualizations of the global embeddings for different tasks in CIL and FIL, where small colored points represent the global embeddings of various classes, and the large points with black borders represent the corresponding class prototypes. 
From Figure \ref{Fig: Vis_CIL}, we observe that as the number of CIL tasks increases, each task includes both the global embeddings of the current task and those generated based on the class prototypes of previous tasks. 
This indicates that V-LETO can maintain the performance of both current and prior tasks simultaneously. 
Figures \ref{Fig: Vis_CIL}(e)-(h) show that as FIL tasks increase (from task 1 to task 4), the number of features grows, and samples of the same class become more tightly clustered, indicating improved model performance.
Therefore, V-LETO effectively addresses both class and feature catastrophic forgetting while efficiently managing both CIL and FIL tasks.

%% file: sec/conclusion.tex
\section{Conclusion}
We propose a novel method, V-LETO,
which mitigates the issue of catastrophic forgetting and enhances model performance by evolving prototype knowledge across tasks. 
We propose a prototype generation method within VFL and leverage a global prototype table to enable knowledge transfer between tasks. 
To address catastrophic forgetting of prior task knowledge, we propose an evolving prototype module based on the PG module. This module integrates prototype knowledge from both previous and current tasks to construct global prototypes, thereby optimizing the global model.
Additionally, we propose a MO module that restricts updates to specific parameters of the local model to mitigate catastrophic forgetting of prior task knowledge.
Extensive experiments demonstrate that our method, V-LETO, outperforms current state-of-the-art methods.

%% file: Main_AAAI2026.bbl
\begin{thebibliography}{40}
\providecommand{\natexlab}[1]{#1}

\bibitem[{Abouelnaga et~al.(2016)Abouelnaga, Ali, Rady, and Moustafa}]{abouelnaga2016cifar}
Abouelnaga, Y.; Ali, O.~S.; Rady, H.; and Moustafa, M. 2016.
\newblock {Cifar-10}: Knn-based ensemble of classifiers.
\newblock In \emph{2016 International Conference on Computational Science and Computational Intelligence}, 1192--1195. Las Vegas, NV, USA.

\bibitem[{Cai and Ma(2022)}]{cai2022theoretical}
Cai, T.~T.; and Ma, R. 2022.
\newblock Theoretical foundations of t-sne for visualizing high-dimensional clustered data.
\newblock \emph{Journal of Machine Learning Research}, 23(301): 1--54.

\bibitem[{Casado et~al.(2023)Casado, Lema, Iglesias, Regueiro, and Barro}]{casado2023ensemble}
Casado, F.~E.; Lema, D.; Iglesias, R.; Regueiro, C.~V.; and Barro, S. 2023.
\newblock Ensemble and continual federated learning for classification tasks.
\newblock \emph{Machine Learning}, 112(9): 3413--3453.

\bibitem[{Castiglia et~al.(2022)Castiglia, Das, Wang, and Patterson}]{castiglia2022compressed}
Castiglia, T.~J.; Das, A.; Wang, S.; and Patterson, S. 2022.
\newblock {Compressed-VFL}: Communication-efficient learning with vertically partitioned data.
\newblock In \emph{International Conference on Machine Learning}, 2738--2766. Baltimore, Maryland, {USA}.

\bibitem[{Feng~Qiang(2022)}]{author2024}
Feng~Qiang, e.~a., Boyan~Wei. 2022.
\newblock White Paper on the Application of Federated Learning Technology in Finance.
\newblock Http://www.hbbill.com/uploadFiles/-16/548/058/54/

\bibitem[{Gao et~al.(2024)Gao, Yang, Yu, Kang, and Li}]{gao2024fedprok}
Gao, X.; Yang, X.; Yu, H.; Kang, Y.; and Li, T. 2024.
\newblock FedProK: Trustworthy Federated Class-Incremental Learning via Prototypical Feature Knowledge Transfer.
\newblock In \emph{Proceedings of the IEEE/CVF Conference on Computer Vision and Pattern Recognition}, 4205--4214. Seattle, WA, USA.

\bibitem[{Hou et~al.(2023)Hou, Gu, Xu, and Qian}]{10227560}
Hou, C.; Gu, S.; Xu, C.; and Qian, Y. 2023.
\newblock Incremental Learning for Simultaneous Augmentation of Feature and Class.
\newblock \emph{IEEE Transactions on Pattern Analysis and Machine Intelligence}, 45(12): 14789--14806.

\bibitem[{Hu et~al.(2019)Hu, Chen, Peng, Yu, Gao, and Hu}]{8410016}
Hu, C.; Chen, Y.; Peng, X.; Yu, H.; Gao, C.; and Hu, L. 2019.
\newblock A Novel Feature Incremental Learning Method for Sensor-Based Activity Recognition.
\newblock \emph{IEEE Transactions on Knowledge and Data Engineering}, 31(6): 1038--1050.

\bibitem[{Krawczyk and Gepperth(2024)}]{krawczyk2024analysis}
Krawczyk, A.; and Gepperth, A. 2024.
\newblock An analysis of best-practice strategies for replay and rehearsal in continual learning.
\newblock In \emph{Proceedings of the IEEE/CVF Conference on Computer Vision and Pattern Recognition}, 4196--4204. Seattle,WA, USA.

\bibitem[{Lang(1995)}]{lang1995newsweeder}
Lang, K. 1995.
\newblock Newsweeder: Learning to filter netnews.
\newblock In \emph{Machine learning proceedings 1995}, 331--339. Elsevier.

\bibitem[{Lebichot et~al.(2024)Lebichot, Siblini, Paldino, Le~Borgne, Obl{\'e}, and Bontempi}]{lebichot2024assessment}
Lebichot, B.; Siblini, W.; Paldino, G.~M.; Le~Borgne, Y.-A.; Obl{\'e}, F.; and Bontempi, G. 2024.
\newblock Assessment of catastrophic forgetting in continual credit card fraud detection.
\newblock \emph{Expert Systems with Applications}, 249(99): 123445.

\bibitem[{Li et~al.(2024{\natexlab{a}})Li, Su, Zhang, and Wang}]{li2024continual}
Li, S.; Su, T.; Zhang, X.; and Wang, Z. 2024{\natexlab{a}}.
\newblock Continual learning with knowledge distillation: A survey.
\newblock \emph{IEEE Transactions on Neural Networks and Learning Systems}, 36(6): 9798 -- 9818.

\bibitem[{Li et~al.(2024{\natexlab{b}})Li, Li, Wang, Li, Zhong, and Zhang}]{li2024towards}
Li, Y.; Li, Q.; Wang, H.; Li, R.; Zhong, W.; and Zhang, G. 2024{\natexlab{b}}.
\newblock Towards Efficient Replay in Federated Incremental Learning.
\newblock In \emph{Proceedings of the IEEE/CVF Conference on Computer Vision and Pattern Recognition}, 12820--12829. Seattle, WA, USA.

\bibitem[{Liao et~al.(2025)Liao, Fu, Zhang, Yang, Chen, Ng, Huang, and Zheng}]{liao2025privacy}
Liao, T.; Fu, L.; Zhang, L.; Yang, L.; Chen, C.; Ng, M.~K.; Huang, H.; and Zheng, Z. 2025.
\newblock Privacy-preserving vertical federated learning with tensor decomposition for data missing features.
\newblock \emph{IEEE Transactions on Information Forensics and Security}, 20: 3445 -- 3460.

\bibitem[{Liu et~al.(2024)Liu, Kang, Zou, Pu, He, Ye, Ouyang, Zhang, and Yang}]{liu2024vertical}
Liu, Y.; Kang, Y.; Zou, T.; Pu, Y.; He, Y.; Ye, X.; Ouyang, Y.; Zhang, Y.-Q.; and Yang, Q. 2024.
\newblock Vertical federated learning: Concepts, advances, and challenges.
\newblock \emph{IEEE Transactions on Knowledge and Data Engineering}, 36(7): 3615 -- 3634.

\bibitem[{Liu et~al.(2023)Liu, Sun, Song, Zhang, Yan, Qiu, Jiang, and Li}]{liu2023vertical}
Liu, Z.; Sun, H.; Song, J.; Zhang, B.; Yan, Y.; Qiu, B.; Jiang, L.; and Li, J. 2023.
\newblock Vertical Federated Learning Architecture for Power Company and Financial Company and Electricity Pricing Model Considering User Credit Evaluation.
\newblock In \emph{2023 3rd International Conference on Consumer Electronics and Computer Engineering (ICCECE)}, 820--826. Guangzhou, China.

\bibitem[{Luo et~al.(2023)Luo, Li, Lan, and Gao}]{luo2023gradma}
Luo, K.; Li, X.; Lan, Y.; and Gao, M. 2023.
\newblock Gradma: A gradient-memory-based accelerated federated learning with alleviated catastrophic forgetting.
\newblock In \emph{Proceedings of the IEEE/CVF Conference on Computer Vision and Pattern Recognition}, 3708--3717. Seattle, WA, USA.

\bibitem[{Luo et~al.(2021{\natexlab{a}})Luo, Chen, Hu, Zhang, Liang, and Feng}]{NEURIPS2021_2f2b2656}
Luo, M.; Chen, F.; Hu, D.; Zhang, Y.; Liang, J.; and Feng, J. 2021{\natexlab{a}}.
\newblock No Fear of Heterogeneity: Classifier Calibration for Federated Learning with Non-IID Data.
\newblock In \emph{Advances in Neural Information Processing Systems}, 5972--5984. Virtual.

\bibitem[{Luo et~al.(2021{\natexlab{b}})Luo, Wu, Xiao, and Ooi}]{luo2021feature}
Luo, X.; Wu, Y.; Xiao, X.; and Ooi, B.~C. 2021{\natexlab{b}}.
\newblock Feature inference attack on model predictions in vertical federated learning.
\newblock In \emph{2021 IEEE 37th International Conference on Data Engineering}, 181--192. Chania, Greece.

\bibitem[{Ma et~al.(2022)Ma, Xie, Wang, Chen, and Shou}]{ma2022continual}
Ma, Y.; Xie, Z.; Wang, J.; Chen, K.; and Shou, L. 2022.
\newblock Continual Federated Learning Based on Knowledge Distillation.
\newblock In \emph{Proceedings of the Thirty-First International Joint Conference on Artificial Intelligence}, 2182--2188. Vienna, Austria.

\bibitem[{Ni et~al.(2024)Ni, Gu, Fan, and Hou}]{ni2024feature}
Ni, H.; Gu, S.; Fan, R.; and Hou, C. 2024.
\newblock Feature incremental learning with causality.
\newblock \emph{Pattern Recognition}, 146(99): 110033.

\bibitem[{Qiao and Mahdavi(2024)}]{qiao2024learn}
Qiao, F.; and Mahdavi, M. 2024.
\newblock Learn more, but bother less: parameter efficient continual learning.
\newblock \emph{Advances in Neural Information Processing Systems}, 37: 97476--97498.

\bibitem[{Qiu et~al.(2024)Qiu, Pu, Liu, Liu, Yue, Zhu, Li, Li, and Ji}]{qiu2024integer}
Qiu, P.; Pu, Y.; Liu, Y.; Liu, W.; Yue, Y.; Zhu, X.; Li, L.; Li, J.; and Ji, S. 2024.
\newblock Integer Is Enough: When Vertical Federated Learning Meets Rounding.
\newblock In \emph{Proceedings of the AAAI Conference on Artificial Intelligence}, 14704--14712. Vancouver, Canada.

\bibitem[{Romanini et~al.(2021)Romanini, Hall, Papadopoulos, Titcombe, Ismail, Cebere, Sandmann, Roehm, and Hoeh}]{romanini2021pyvertical}
Romanini, D.; Hall, A.~J.; Papadopoulos, P.; Titcombe, T.; Ismail, A.; Cebere, T.; Sandmann, R.; Roehm, R.; and Hoeh, M.~A. 2021.
\newblock Pyvertical: A vertical federated learning framework for multi-headed splitnn.
\newblock \emph{arXiv preprint arXiv:2104.00489}, PP(99): 1--9.

\bibitem[{Sakib and Das(2024)}]{sakib2024explainable}
Sakib, S.~K.; and Das, A.~B. 2024.
\newblock Explainable Vertical Federated Learning for Healthcare: Ensuring Privacy and Optimal Accuracy.
\newblock In \emph{2024 IEEE International Conference on Big Data}, 5068--5077. Washington, DC, USA.

\bibitem[{Shenaj et~al.(2023)Shenaj, Toldo, Rigon, and Zanuttigh}]{shenaj2023asynchronous}
Shenaj, D.; Toldo, M.; Rigon, A.; and Zanuttigh, P. 2023.
\newblock Asynchronous federated continual learning.
\newblock In \emph{Proceedings of the IEEE/CVF Conference on Computer Vision and Pattern Recognition}, 5055--5063. Vancouver, BC, Canada.

\bibitem[{Wang et~al.(2023)Wang, Gu, Zhang, Li, Wang, and Ling}]{wang2024unified}
Wang, G.; Gu, B.; Zhang, Q.; Li, X.; Wang, B.; and Ling, C.~X. 2023.
\newblock A unified solution for privacy and communication efficiency in vertical federated learning.
\newblock In \emph{Advances in Neural Information Processing Systems}, 13480--13491. New Orleans, LA, USA.

\bibitem[{Wang et~al.(2025{\natexlab{a}})Wang, Zhang, Huang, Gai, Wang, and Shen}]{wang2025ropa}
Wang, L.; Zhang, Z.; Huang, M.; Gai, K.; Wang, J.; and Shen, Y. 2025{\natexlab{a}}.
\newblock RoPA: Robust Privacy-Preserving Forward Aggregation for Split Vertical Federated Learning.
\newblock \emph{IEEE Transactions on Network and Service Management}, 1.

\bibitem[{Wang, Liu, and Li(2024)}]{wang2024traceable}
Wang, Q.; Liu, B.; and Li, Y. 2024.
\newblock Traceable Federated Continual Learning.
\newblock In \emph{Proceedings of the IEEE/CVF Conference on Computer Vision and Pattern Recognition}, 12872--12881. Seattle, WA, USA.

\bibitem[{Wang et~al.(2025{\natexlab{b}})Wang, Gai, Yu, Zhang, and Zhu}]{wang2023bdvfl}
Wang, S.; Gai, K.; Yu, J.; Zhang, Z.; and Zhu, L. 2025{\natexlab{b}}.
\newblock Pravfed: Practical heterogeneous vertical federated learning via representation learning.
\newblock \emph{IEEE Transactions on Information Forensics and Security}, 20(1): 2693 -- 2705.

\bibitem[{Wang et~al.(2025{\natexlab{c}})Wang, Gai, Yu, Zhang, and Zhu}]{wang2025pravfed}
Wang, S.; Gai, K.; Yu, J.; Zhang, Z.; and Zhu, L. 2025{\natexlab{c}}.
\newblock PraVFed: Practical Heterogeneous Vertical Federated Learning via Representation Learning.
\newblock \emph{IEEE Transactions on Information Forensics and Security}, PP(99): 1.

\bibitem[{Xiao, Rasul, and Vollgraf(2017)}]{xiao2017fashion}
Xiao, H.; Rasul, K.; and Vollgraf, R. 2017.
\newblock Fashion-mnist: a novel image dataset for benchmarking machine learning algorithms.
\newblock \emph{arXiv preprint arXiv:1708.07747}, PP(99): 1.

\bibitem[{Yang, Huang, and Ye(2023)}]{yang2023dynamic}
Yang, X.; Huang, W.; and Ye, M. 2023.
\newblock Dynamic Personalized Federated Learning with Adaptive Differential Privacy.
\newblock In \emph{Advances in Neural Information Processing Systems}, 72181--72192. New Orleans, LA, USA.

\bibitem[{Yang et~al.(2024)Yang, Yu, Gao, Wang, Zhang, and Li}]{yang2024federated}
Yang, X.; Yu, H.; Gao, X.; Wang, H.; Zhang, J.; and Li, T. 2024.
\newblock Federated continual learning via knowledge fusion: A survey.
\newblock \emph{IEEE Transactions on Knowledge and Data Engineering}, 38(8): 3832--3850.

\bibitem[{Ye et~al.(2025)Ye, Shen, Du, Snezhko, Kovalev, and Yuen}]{ye2025vertical}
Ye, M.; Shen, W.; Du, B.; Snezhko, E.; Kovalev, V.; and Yuen, P.~C. 2025.
\newblock Vertical federated learning for effectiveness, security, applicability: A survey.
\newblock \emph{ACM Computing Surveys}, 57(9): 1--32.

\bibitem[{Yoon et~al.(2021)Yoon, Jeong, Lee, Yang, and Hwang}]{yoon2021federated}
Yoon, J.; Jeong, W.; Lee, G.; Yang, E.; and Hwang, S.~J. 2021.
\newblock Federated continual learning with weighted inter-client transfer.
\newblock In \emph{International Conference on Machine Learning}, 12073--12086. Virtual Event.

\bibitem[{Yu et~al.(2024)Yu, Yang, Gao, Feng, Wang, Kang, and Li}]{yu2024overcoming}
Yu, H.; Yang, X.; Gao, X.; Feng, Y.; Wang, H.; Kang, Y.; and Li, T. 2024.
\newblock Overcoming Spatial-Temporal Catastrophic Forgetting for Federated Class-Incremental Learning.
\newblock In \emph{ACM Multimedia 2024}, 1--9. Melbourne, Australia.

\bibitem[{Zhang et~al.(2022{\natexlab{a}})Zhang, Guo, Qu, Zeng, Wang, Liu, and Zomaya}]{zhang2022adaptive}
Zhang, J.; Guo, S.; Qu, Z.; Zeng, D.; Wang, H.; Liu, Q.; and Zomaya, A.~Y. 2022{\natexlab{a}}.
\newblock Adaptive vertical federated learning on unbalanced features.
\newblock \emph{IEEE Transactions on Parallel and Distributed Systems}, 33(12): 4006--4018.

\bibitem[{Zhang et~al.(2022{\natexlab{b}})Zhang, Guo, Sun, Liu, and Yu}]{zhang2022cross}
Zhang, Z.; Guo, B.; Sun, W.; Liu, Y.; and Yu, Z. 2022{\natexlab{b}}.
\newblock {Cross-FCL}: Toward a cross-edge federated continual learning framework in mobile edge computing systems.
\newblock \emph{IEEE Transactions on Mobile Computing}, 23(1): 313--326.

\bibitem[{Zhu et~al.(2021)Zhu, Zhang, Wang, Yin, and Liu}]{zhu2021prototype}
Zhu, F.; Zhang, X.-Y.; Wang, C.; Yin, F.; and Liu, C.-L. 2021.
\newblock Prototype augmentation and self-supervision for incremental learning.
\newblock In \emph{Proceedings of the IEEE/CVF Conference on Computer Vision and Pattern Recognition}, 5871--5880. Virtual.

\end{thebibliography}
